%% file: acl_latex.tex
\documentclass[11pt]{article}

\usepackage[final]{acl}

\usepackage{times}
\usepackage[utf8]{inputenc}

\usepackage{makecell}
\usepackage{tabularx}

\usepackage{latexsym}

\usepackage[T1]{fontenc}

\usepackage{microtype}

\usepackage{inconsolata}

\usepackage{graphicx}
\usepackage[dvipsnames,table,xcdraw]{xcolor}
\usepackage{amsmath, amssymb}
\usepackage{tcolorbox}
\tcbuselibrary{listings,skins,breakable}

\usepackage{lipsum}               
\tcbuselibrary{breakable}
\usepackage{cuted}
\usepackage{fancyvrb}
\usepackage{listings}
\usepackage{booktabs}
\usepackage{multirow}
\usepackage{geometry}
\usepackage{subcaption}
\usepackage{float}
\usepackage{rotating}
\usepackage{wrapfig}
\usepackage{pifont}
\usepackage{placeins}
\newcommand{\cmark}{\ding{51}}
\newcommand{\xmark}{\ding{55}}
\tcbuselibrary{breakable, skins}
\input{scripts/prompt_defs}

\newtcblisting{tcbverbatim}{
    listing only,
    listing options={basicstyle=\ttfamily},
    colback=gray!10,
    colframe=gray!50,
    breakable
}

\newtcolorbox{examplebox}{
  enhanced,
  breakable,
  colback=gray!3,
  colframe=black,
  boxrule=0.8pt,
  arc=2pt,
  left=6pt,
  right=6pt,
  top=6pt,
  bottom=6pt
}

\title{Is Next-Chunk Reasoning RL Really Better than SFT? \\ Revisiting Training Strategies under no-CoT Data}

\author{
    \textbf{Yinhao Tang}$^{1,2}$\textsuperscript{*} \quad
    \textbf{Youqing Fang}$^{1,2}$\textsuperscript{*} \quad
    \textbf{Yanan Sun}$^{2}$\textsuperscript{\dag} \quad 
    \textbf{Jiangning Liu}$^{2}$\quad \\
    \textbf{Ziyi Wang}$^{2}$\quad
    \textbf{Xun Zhao}$^{2}$\quad
    \textbf{Weiming Zhang}$^{1}$\quad 
    \textbf{Bin Liu}$^{1}$ \quad \\
    \textbf{Kuikun Liu}$^{2}$ \quad
    \textbf{Wenwei Zhang}$^{2}$ \quad
    \textbf{Kai Chen}$^{2}$\ \quad \\
    \textsuperscript{1}University of Science and Technology of China \textsuperscript{2}Shanghai AI Laboratory \\
    \normalsize
    \normalfont
    \texttt{\{tangyinhao,fangyq\}@mail.ustc.edu.cn}, 
    \texttt{\{sunyanan\}@pjlab.org.cn}
}

\begin{document}
\maketitle

\begingroup
\renewcommand{\thefootnote}{}
\footnotetext{* Equal contribution.\, $\dag$ Corresponding author.}
\endgroup

\input{sec/0_abstract}
\input{sec/1_introduction}
\input{sec/3_experimental_setup}
\input{sec/4_main_results}

\input{sec/5_analysis}
\input{sec/2_related}
\input{sec/6_conclusion}

\bibliography{custom}

\appendix



\input{sec/7_appendix}

\end{document}

%% file: scripts/prompt_defs.tex
\definecolor{prompt}{RGB}{225, 240, 248}
\definecolor{prompt-frame}{RGB}{160, 200, 225}
\definecolor{prompt2}{RGB}{225, 240, 248}
\definecolor{prompt2-frame}{RGB}{160, 200, 225}
\definecolor{prompt3}{RGB}{225, 240, 248}
\definecolor{prompt3-frame}{RGB}{160, 200, 225}
\definecolor{prompt4}{RGB}{225, 240, 248}
\definecolor{prompt4-frame}{RGB}{160, 200, 225}
\definecolor{prompt5}{RGB}{225, 240, 248}
\definecolor{prompt5-frame}{RGB}{160, 200, 225}
\usepackage{fontawesome5}
\tcbset{
    prompt_func/.style={
        enhanced,
        breakable,
        colback=prompt,        
        colframe=prompt-frame,            
        fontupper=\normalsize,               
        boxrule=1pt,                       
        arc=4mm,                           
        left=1mm, right=1mm, top=1mm, bottom=1mm, 
        boxsep=4pt,                        
        before skip=10pt, after skip=10pt, 
        overlay={
            \node[anchor=north west, xshift=4pt, text=white] at (frame.north west) {\faDatabase};
        },
        title={~~~~~~~\textbf{Prompt used with GPT-4o to generate annotation for Element Grounding with functional setting}},
        coltitle=white,
        fonttitle=\bfseries\small
    }
}

\tcbset{
    full_content_generation/.style={
        enhanced,
        breakable,
        colback=prompt,        
        colframe=prompt-frame,            
        fontupper=\normalsize,               
        boxrule=1pt,                       
        arc=4mm,                           
        left=1mm, right=1mm, top=1mm, bottom=1mm, 
        boxsep=4pt,                        
        before skip=10pt, after skip=10pt, 
        title={\faDatabase~~~\textbf{Prompt: Full Content Generation}},
        coltitle=white,
        fonttitle=\bfseries\small
    }
}
\tcbset{
    filter_figures/.style={
        enhanced,
        breakable,
        colback=prompt,        
        colframe=prompt-frame,            
        fontupper=\normalsize,               
        boxrule=1pt,                       
        arc=4mm,                           
        left=1mm, right=1mm, top=1mm, bottom=1mm, 
        boxsep=4pt,                        
        before skip=10pt, after skip=10pt, 
        title={\faDatabase~~~\textbf{Prompt: Filter Figures}},
        coltitle=white,
        fonttitle=\bfseries\small
    }
}
\tcbset{
    section_generation/.style={
        enhanced,
        breakable,
        colback=prompt,        
        colframe=prompt-frame,            
        fontupper=\normalsize,               
        boxrule=1pt,                       
        arc=4mm,                           
        left=1mm, right=1mm, top=1mm, bottom=1mm, 
        boxsep=4pt,                        
        before skip=10pt, after skip=10pt, 
        title={\faDatabase~~~\textbf{Prompt: Section Generation}},
        coltitle=white,
        fonttitle=\bfseries\small
    }
}
\tcbset{
    html_generation/.style={
        enhanced,
        breakable,
        colback=prompt,        
        colframe=prompt-frame,            
        fontupper=\normalsize,               
        boxrule=1pt,                       
        arc=4mm,                           
        left=1mm, right=1mm, top=1mm, bottom=1mm, 
        boxsep=4pt,                        
        before skip=10pt, after skip=10pt, 
        title={\faDatabase~~~\textbf{Prompt: HTML Generation}},
        coltitle=white,
        fonttitle=\bfseries\small
    }
}
\tcbset{
    full_content_review/.style={
        enhanced,
        breakable,
        colback=prompt,        
        colframe=prompt-frame,            
        fontupper=\normalsize,               
        boxrule=1pt,                       
        arc=4mm,                           
        left=1mm, right=1mm, top=1mm, bottom=1mm, 
        boxsep=4pt,                        
        before skip=10pt, after skip=10pt, 
        title={\faDatabase~~~\textbf{Prompt: Full Content Review}},
        coltitle=white,
        fonttitle=\bfseries\small
    }
}
\tcbset{
    full_content_revise/.style={
        enhanced,
        breakable,
        colback=prompt,        
        colframe=prompt-frame,            
        fontupper=\normalsize,               
        boxrule=1pt,                       
        arc=4mm,                           
        left=1mm, right=1mm, top=1mm, bottom=1mm, 
        boxsep=4pt,                        
        before skip=10pt, after skip=10pt, 
        title={\faDatabase~~~\textbf{Prompt: Full Content Revise}},
        coltitle=white,
        fonttitle=\bfseries\small
    }
}
\tcbset{
    text_content_generation/.style={
        enhanced,
        breakable,
        colback=prompt,        
        colframe=prompt-frame,            
        fontupper=\normalsize,               
        boxrule=1pt,                       
        arc=4mm,                           
        left=1mm, right=1mm, top=1mm, bottom=1mm, 
        boxsep=4pt,                        
        before skip=10pt, after skip=10pt, 
        title={\faDatabase~~~\textbf{Prompt: Text Content Generation}},
        coltitle=white,
        fonttitle=\bfseries\small
    }
}
\tcbset{
    html_review/.style={
        enhanced,
        breakable,
        colback=prompt,        
        colframe=prompt-frame,            
        fontupper=\normalsize,               
        boxrule=1pt,                       
        arc=4mm,                           
        left=1mm, right=1mm, top=1mm, bottom=1mm, 
        boxsep=4pt,                        
        before skip=10pt, after skip=10pt, 
        title={\faDatabase~~~\textbf{Prompt: HTML Review}},
        coltitle=white,
        fonttitle=\bfseries\small
    }
}
\tcbset{
    html_revise/.style={
        enhanced,
        breakable,
        colback=prompt,        
        colframe=prompt-frame,            
        fontupper=\normalsize,               
        boxrule=1pt,                       
        arc=4mm,                           
        left=1mm, right=1mm, top=1mm, bottom=1mm, 
        boxsep=4pt,                        
        before skip=10pt, after skip=10pt, 
        title={\faDatabase~~~\textbf{Prompt: HTML Revise}},
        coltitle=white,
        fonttitle=\bfseries\small
    }
}
\tcbset{
    prompt_aethetic_element/.style={
        enhanced,
        breakable,
        colback=prompt,        
        colframe=prompt-frame,            
        fontupper=\normalsize,               
        boxrule=1pt,                       
        arc=4mm,                           
        left=1mm, right=1mm, top=1mm, bottom=1mm, 
        boxsep=4pt,                        
        before skip=10pt, after skip=10pt, 
        title={\faDatabase~~~\textbf{Prompt: Element Quality Judge}},
        coltitle=white,
        fonttitle=\bfseries\small
    }
}

\tcbset{
    prompt_aethetic_Layout/.style={
        enhanced,
        breakable,
        colback=prompt,        
        colframe=prompt-frame,            
        fontupper=\normalsize,               
        boxrule=1pt,                       
        arc=4mm,                           
        left=1mm, right=1mm, top=1mm, bottom=1mm, 
        boxsep=4pt,                        
        before skip=10pt, after skip=10pt, 
        title={\faDatabase~~~\textbf{Prompt: Layout Balance Judge}},
        coltitle=white,
        fonttitle=\bfseries\small
    }
}

\tcbset{prompt_aethetic_engagement/.style={
        enhanced,
        breakable,
        colback=prompt,        
        colframe=prompt-frame,            
        fontupper=\normalsize,               
        boxrule=1pt,                       
        arc=4mm,                           
        left=1mm, right=1mm, top=1mm, bottom=1mm, 
        boxsep=4pt,                        
        before skip=10pt, after skip=10pt, 
        title={\faDatabase~~~\textbf{Prompt: Engagement Judge}},
        coltitle=white,
        fonttitle=\bfseries\small
    }
}

\tcbset{prompt_info_clarity/.style={
        enhanced,
        breakable,
        colback=prompt,        
        colframe=prompt-frame,            
        fontupper=\normalsize,               
        boxrule=1pt,                       
        arc=4mm,                           
        left=1mm, right=1mm, top=1mm, bottom=1mm, 
        boxsep=4pt,                        
        before skip=10pt, after skip=10pt, 
        title={\faDatabase~~~\textbf{Prompt: Clarity Judge}},
        coltitle=white,
        fonttitle=\bfseries\small
    }
}

\tcbset{prompt_info_content/.style={
        enhanced,
        breakable,
        colback=prompt,        
        colframe=prompt-frame,            
        fontupper=\normalsize,               
        boxrule=1pt,                       
        arc=4mm,                           
        left=1mm, right=1mm, top=1mm, bottom=1mm, 
        boxsep=4pt,                        
        before skip=10pt, after skip=10pt, 
        title={\faDatabase~~~\textbf{Prompt: Content Completeness Judge}},
        coltitle=white,
        fonttitle=\bfseries\small
    }
}

\tcbset{prompt_info_logic/.style={
        enhanced,
        breakable,
        colback=prompt,        
        colframe=prompt-frame,            
        fontupper=\normalsize,               
        boxrule=1pt,                       
        arc=4mm,                           
        left=1mm, right=1mm, top=1mm, bottom=1mm, 
        boxsep=4pt,                        
        before skip=10pt, after skip=10pt, 
        title={\faDatabase~~~\textbf{Prompt: Logical Flow Judge}},
        coltitle=white,
        fonttitle=\bfseries\small
    }
}

\tcbset{prompt_verbatim_question/.style={
        enhanced,
        breakable,
        colback=prompt,        
        colframe=prompt-frame,            
        fontupper=\normalsize,               
        boxrule=1pt,                       
        arc=4mm,                           
        left=1mm, right=1mm, top=1mm, bottom=1mm, 
        boxsep=4pt,                        
        before skip=10pt, after skip=10pt, 
        title={\faDatabase~~~\textbf{Prompt: Generate Verbatim QA}},
        coltitle=white,
        fonttitle=\bfseries\small
    }
}

\tcbset{prompt_interpretive_question/.style={
        enhanced,
        breakable,
        colback=prompt,        
        colframe=prompt-frame,            
        fontupper=\normalsize,               
        boxrule=1pt,                       
        arc=4mm,                           
        left=1mm, right=1mm, top=1mm, bottom=1mm, 
        boxsep=4pt,                        
        before skip=10pt, after skip=10pt, 
        title={\faDatabase~~~\textbf{Prompt: Generate Interpretive QA}},
        coltitle=white,
        fonttitle=\bfseries\small
    }
}

\tcbset{prompt_answer_agent/.style={
        enhanced,
        breakable,
        colback=prompt,        
        colframe=prompt-frame,            
        fontupper=\normalsize,               
        boxrule=1pt,                       
        arc=4mm,                           
        left=1mm, right=1mm, top=1mm, bottom=1mm, 
        boxsep=4pt,                        
        before skip=10pt, after skip=10pt, 
        title={\faDatabase~~~\textbf{Prompt: Answer Questions}},
        coltitle=white,
        fonttitle=\bfseries\small
    }
}

\tcbset{prompt_4o_image/.style={
        enhanced,
        breakable,
        colback=prompt,        
        colframe=prompt-frame,            
        fontupper=\normalsize,               
        boxrule=1pt,                       
        arc=4mm,                           
        left=1mm, right=1mm, top=1mm, bottom=1mm, 
        boxsep=4pt,                        
        before skip=10pt, after skip=10pt, 
        title={\faDatabase~~~\textbf{Prompt: \texttt{4o-Image}}},
        coltitle=white,
        fonttitle=\bfseries\small
    }
}

\tcbset{prompt_owl_4o/.style={
        enhanced,
        breakable,
        colback=prompt,        
        colframe=prompt-frame,            
        fontupper=\normalsize,               
        boxrule=1pt,                       
        arc=4mm,                           
        left=1mm, right=1mm, top=1mm, bottom=1mm, 
        boxsep=4pt,                        
        before skip=10pt, after skip=10pt, 
        title={\faDatabase~~~\textbf{Prompt: \texttt{OWL-4o}}},
        coltitle=white,
        fonttitle=\bfseries\small
    }
}

\tcbset{prompt_4o_html/.style={
        enhanced,
        breakable,
        colback=prompt,        
        colframe=prompt-frame,            
        fontupper=\normalsize,               
        boxrule=1pt,                       
        arc=4mm,                           
        left=1mm, right=1mm, top=1mm, bottom=1mm, 
        boxsep=4pt,                        
        before skip=10pt, after skip=10pt, 
        title={\faDatabase~~~\textbf{Prompt: \texttt{4o-HTML}}},
        coltitle=white,
        fonttitle=\bfseries\small
    }
}

\tcbset{prompt_parser_summarizer/.style={
        enhanced,
        breakable,
        colback=prompt,        
        colframe=prompt-frame,            
        fontupper=\normalsize,               
        boxrule=1pt,                       
        arc=4mm,                           
        left=1mm, right=1mm, top=1mm, bottom=1mm, 
        boxsep=4pt,                        
        before skip=10pt, after skip=10pt, 
        title={\faDatabase~~~\textbf{Prompt: Paper Summarizer}},
        coltitle=white,
        fonttitle=\bfseries\small
    }
}

\tcbset{prompt_parser_filter/.style={
        enhanced,
        breakable,
        colback=prompt,        
        colframe=prompt-frame,            
        fontupper=\normalsize,               
        boxrule=1pt,                       
        arc=4mm,                           
        left=1mm, right=1mm, top=1mm, bottom=1mm, 
        boxsep=4pt,                        
        before skip=10pt, after skip=10pt, 
        title={\faDatabase~~~\textbf{Prompt: Figure Filter}},
        coltitle=white,
        fonttitle=\bfseries\small
    }
}

\tcbset{prompt_planner_matching/.style={
        enhanced,
        breakable,
        colback=prompt,        
        colframe=prompt-frame,            
        fontupper=\normalsize,               
        boxrule=1pt,                       
        arc=4mm,                           
        left=1mm, right=1mm, top=1mm, bottom=1mm, 
        boxsep=4pt,                        
        before skip=10pt, after skip=10pt, 
        title={\faDatabase~~~\textbf{Prompt: Asset Matching}},
        coltitle=white,
        fonttitle=\bfseries\small
    }
}

\tcbset{prompt_planner_painter/.style={
        enhanced,
        breakable,
        colback=prompt,        
        colframe=prompt-frame,            
        fontupper=\normalsize,               
        boxrule=1pt,                       
        arc=4mm,                           
        left=1mm, right=1mm, top=1mm, bottom=1mm, 
        boxsep=4pt,                        
        before skip=10pt, after skip=10pt, 
        title={\faDatabase~~~\textbf{Prompt: Painter}},
        coltitle=white,
        fonttitle=\bfseries\small
    }
}

\tcbset{prompt_planner_commenter/.style={
        enhanced,
        breakable,
        colback=prompt,        
        colframe=prompt-frame,            
        fontupper=\normalsize,               
        boxrule=1pt,                       
        arc=4mm,                           
        left=1mm, right=1mm, top=1mm, bottom=1mm, 
        boxsep=4pt,                        
        before skip=10pt, after skip=10pt, 
        title={\faDatabase~~~\textbf{Prompt: Commenter}},
        coltitle=white,
        fonttitle=\bfseries\small
    }
}

\tcbset{
    vlm_aesthetics_judge/.style={
        enhanced,
        breakable,
        colback=prompt,        
        colframe=prompt-frame,            
        fontupper=\normalsize,               
        boxrule=1pt,                       
        arc=4mm,                           
        left=1mm, right=1mm, top=1mm, bottom=1mm, 
        boxsep=4pt,                        
        before skip=10pt, after skip=10pt, 
        title={\faDatabase~~~\textbf{Prompt: Aesthetics Quality Judge}},
        coltitle=white,
        fonttitle=\bfseries\small
    }
}
\tcbset{
    vlm_element_judge/.style={
        enhanced,
        breakable,
        colback=prompt,        
        colframe=prompt-frame,            
        fontupper=\normalsize,               
        boxrule=1pt,                       
        arc=4mm,                           
        left=1mm, right=1mm, top=1mm, bottom=1mm, 
        boxsep=4pt,                        
        before skip=10pt, after skip=10pt, 
        title={\faDatabase~~~\textbf{Prompt: Element Quality Judge}},
        coltitle=white,
        fonttitle=\bfseries\small
    }
}

\tcbset{
    vlm_layout_judge/.style={
        enhanced,
        breakable,
        colback=prompt,        
        colframe=prompt-frame,            
        fontupper=\normalsize,               
        boxrule=1pt,                       
        arc=4mm,                           
        left=1mm, right=1mm, top=1mm, bottom=1mm, 
        boxsep=4pt,                        
        before skip=10pt, after skip=10pt, 
        title={\faDatabase~~~\textbf{Prompt: Layout Quality Judge}},
        coltitle=white,
        fonttitle=\bfseries\small
    }
}

\tcbset{
    rpt_prompt/.style={
        enhanced,
        breakable,
        colback=prompt,
        colframe=prompt-frame,
        fontupper=\normalsize,
        boxrule=1pt,
        arc=4mm,
        left=1mm, right=1mm, top=1mm, bottom=1mm,
        boxsep=4pt,
        before skip=10pt, after skip=10pt,
        title={\faDatabase~~~\textbf{Prompt: RPT (Token-Level Reconstruction)}},
        coltitle=black,
        fonttitle=\bfseries\small
    }
}

\tcbset{
    rlpt_prompt/.style={
        enhanced,
        breakable,
        colback=prompt,
        colframe=prompt-frame,
        fontupper=\normalsize,
        boxrule=1pt,
        arc=4mm,
        left=1mm, right=1mm, top=1mm, bottom=1mm,
        boxsep=4pt,
        before skip=10pt, after skip=10pt,
        title={\faDatabase~~~\textbf{Prompt: RLPT (Sentence-Level Reconstruction)}},
        coltitle=black,
        fonttitle=\bfseries\small
    }
}

\tcbset{
    rlvr_prompt/.style={
        enhanced,
        breakable,
        colback=prompt,
        colframe=prompt-frame,
        fontupper=\normalsize,
        boxrule=1pt,
        arc=4mm,
        left=1mm, right=1mm, top=1mm, bottom=1mm,
        boxsep=4pt,
        before skip=10pt, after skip=10pt,
        title={\faDatabase~~~\textbf{Prompt: RLVR}},
        coltitle=black,
        fonttitle=\bfseries\small
    }
}

%% file: sec/0_abstract.tex
\begin{abstract}
Recent work proposes \textit{next-chunk reasoning RL} for leveraging \textit{no-CoT data}---corpora such as worked solutions and textbook derivations that contain reasoning-rich content but lack explicit chain-of-thought annotations. The method trains a model to generate implicit reasoning traces and rewards them by their ability to predict the next chunk of text. While promising, existing evaluations primarily compare against conventional SFT baselines, leaving open whether the gains come from the RL formulation itself or from more effectively exposing the model to no-CoT data. We address this question with a controlled study of next-chunk reasoning RL and a simple but previously overlooked alternative: \textit{Mixed SFT}, a single supervised fine-tuning stage that jointly trains on no-CoT and long-CoT data. Despite its simplicity, Mixed SFT achieves a clearly higher post-RLVR performance ceiling than next-chunk reasoning RL while requiring over 60$\times$ less training compute. The advantage is consistent across in-domain mathematical reasoning and out-of-domain reasoning tasks. Moreover, we show that higher pre-RLVR accuracy does not necessarily translate into higher post-RLVR accuracy, highlighting the need to evaluate no-CoT training strategies in the context of the full post-training pipeline.
\end{abstract}

%% file: sec/1_introduction.tex
\begin{figure*}[t]
    \centering
    \includegraphics[width=1\textwidth]{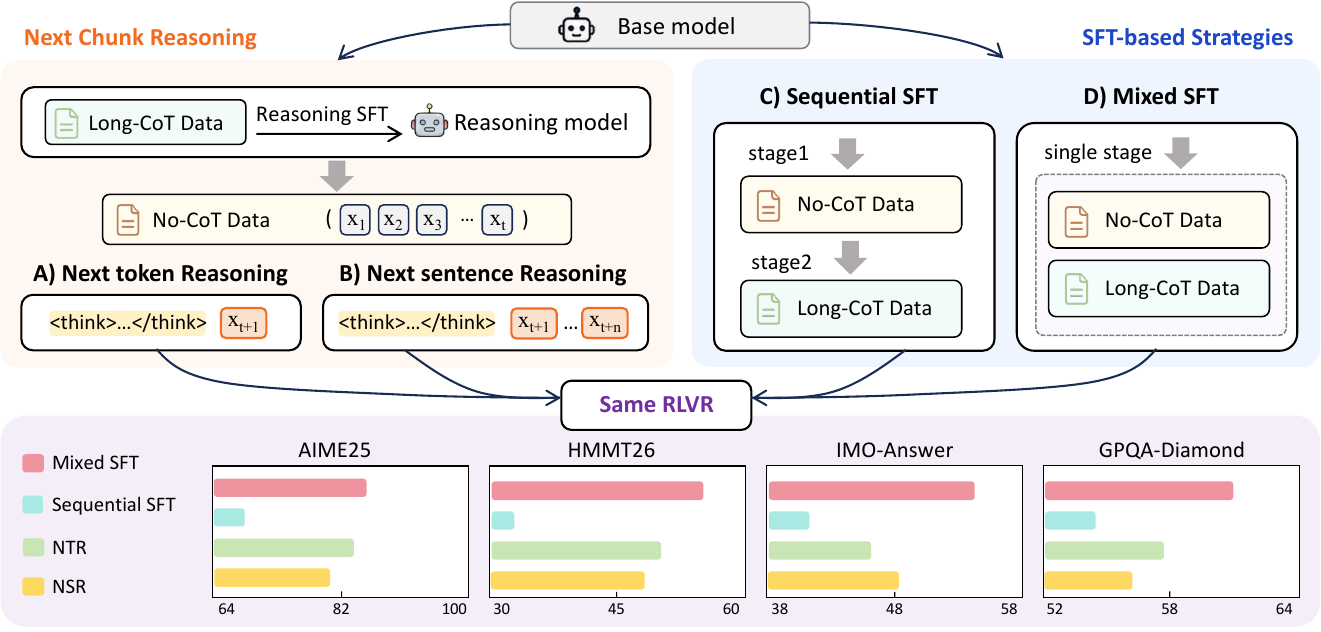}
    \caption{
    Compared strategies for leveraging no-CoT data (\textbf{top}) and their post-RLVR performance (\textbf{bottom}). NCR applies RL rewards over predicted tokens (A: NTR) or sentences (B: NSR). Sequential SFT (C) trains on no-CoT and long-CoT data in two stages, while Mixed SFT (D) combines both in a single stage. All strategies share the same RLVR stage; bar charts show post-RLVR accuracy on four representative benchmarks, with Mixed SFT consistently achieving the highest scores.
    }
    \label{fig:teaser}
\end{figure*}

\section{Introduction}

Recent progress in reasoning-oriented post-training has been largely driven by training on long chain-of-thought (CoT) demonstrations~\citep{wei2022chain, jaech2024openai, guo2025deepseek, agarwal2025gpt, bai2026intern}, which are typically obtained by rejection-sampling correct trajectories from a strong reasoning teacher, making them expensive to scale. In contrast, a far larger and more readily available portion of text, such as worked solutions, textbook derivations, and research papers, presents only conclusions or compressed explanations. These \textit{no-CoT data} lack explicit reasoning traces but still carry the knowledge and solution patterns that reasoning models need to acquire, raising a central question for scaling reasoning post-training~\citep{lambert2024tulu, zhang2025survey}: how should they be turned into useful training signal?

A direct approach is to perform supervised fine-tuning (SFT) on no-CoT data~\citep{wei2021finetuned, zhou2023lima}. Yet no-CoT data lacks explicit long-CoT traces, so naive SFT may distort the model's existing reasoning format~\citep{chu2025sft, matsutani2025rl, fang2026mindcopilot}. In response, recent works turn to reinforcement learning (RL), specifically \textit{next-chunk reasoning} (NCR), in which the model generates implicit reasoning and is rewarded for how well that reasoning predicts the next chunk of the corpus. Within NCR, two variants exist by prediction granularity: \textit{next-token reasoning} (NTR) rewards prediction of the next token (e.g., RPT~\citep{dong2025reinforcement}, RLP~\citep{hatamizadeh2025rlp}, RMT~\citep{tian2025reinforcement}), while \textit{next-sentence reasoning} (NSR) rewards future sentences or text spans (e.g., RLPT~\citep{li2025reinforcement}, PretrainZero~\citep{xing2025pretrainzero}). Both observe gains over SFT trained on the same no-CoT data, albeit at substantially higher training cost since each update step requires online rollouts, reward computation, and policy optimization.

While these reported gains are encouraging, they \textbf{leave open whether the gains stem from the RL formulation itself or from more effectively exposing the model to no-CoT data}. The existing comparisons primarily evaluate next-chunk reasoning RL against SFT baselines trained on no-CoT data alone, hereafter no-CoT SFT, which is not a reasonable reference for the subsequent RLVR stage: although several of these works initialize from a reasoning rather than a base model (Tab.~\ref{tab:cot_recon_comparison}), training on no-CoT data alone still disrupts the long-CoT format the model relies on, leaving the post-SFT checkpoint unable to produce the structured reasoning that RLVR builds upon. To preserve the long-CoT format while still injecting no-CoT knowledge, a natural alternative is \textit{Mixed SFT}, a single stage that jointly trains on no-CoT and long-CoT data (Tab.~\ref{tab:cot_recon_comparison}); this is precisely the baseline we introduce.
With Mixed SFT in place, we put the comparison to a direct test and ask: under the same no-CoT data and reinforcement learning with verifiable rewards (RLVR) budget, is next-chunk reasoning really more effective than SFT?

\begin{table}[t]
\centering
\footnotesize
\setlength{\tabcolsep}{3pt}
\caption{
Summary of next-chunk reasoning RL methods, reporting their starting model (base or reasoning), whether they compare against a no-CoT SFT baseline, and whether they compare against Mixed SFT. RPT~\citep{dong2025reinforcement}, RLP~\citep{hatamizadeh2025rlp}, RMT~\citep{tian2025reinforcement}, RLPT~\citep{li2025reinforcement}, PretrainZero~\citep{xing2025pretrainzero}.
}
\label{tab:cot_recon_comparison}
{
\begin{tabular}{llcc}
\toprule
Method & \makecell{Start Model} & \makecell{No-CoT SFT} & \makecell{Mixed SFT} \\
\midrule
\rowcolor{gray!15} \multicolumn{4}{l}{\textit{Next-token Reasoning}} \\
RPT          & reasoning & \cmark & \xmark \\
RLP          & base & \cmark & \xmark \\
RMT          & reasoning & \cmark & \xmark \\
\rowcolor{gray!15} \multicolumn{4}{l}{\textit{Next-sentence Reasoning}} \\
RLPT         & reasoning & \xmark & \xmark \\
PretrainZero & base & \cmark & \xmark \\
\bottomrule
\end{tabular}
}
\end{table}

To answer this question, we move both next-chunk reasoning RL and SFT before RLVR and compare them from the same base model, avoiding confounds from prior post-training. In doing so, we also uncover a systematic bias in intermediate-checkpoint evaluation and show that how no-CoT data is organized within SFT matters as much as whether it is included.

Concretely, as illustrated in Fig.~\ref{fig:teaser}, we compare two paradigms from the same pre-trained base model: (i) next-chunk reasoning RL at different granularities, and (ii) two SFT strategies---Sequential SFT, which trains on no-CoT data first and then on long-CoT data to restore the long-CoT format, and Mixed SFT, which trains on both jointly in a single stage. Beyond the performance comparison, we conduct a mechanistic analysis to understand \textit{why} NCR fails to outperform Mixed SFT: we show that NTR's entropy filter does not select genuinely reasoning-hard tokens, that the generated traces degenerate into local completion, and that suppressing this collapse does not raise the post-RLVR ceiling. We further trace Mixed SFT's pre-RLVR drop to a transient format mismatch that RLVR repairs.
Together, these experiments and analyses lead to the following contributions:

\begin{itemize}

    \item \textbf{Conceptually, we identify an overlooked gap in recent NCR comparisons and introduce Mixed SFT as the missing baseline.} Prior works compare NCR against SFT trained on no-CoT data alone, omitting the formulation that jointly trains on no-CoT and long-CoT data in a single stage.

    \item \textbf{Empirically, we show that Mixed SFT is simpler, more effective, and substantially cheaper than next-chunk reasoning RL for leveraging no-CoT data.} On the same no-CoT data, Mixed SFT reaches a clearly higher post-RLVR ceiling on both in-domain math and out-of-domain reasoning while requiring over 60$\times$ less training compute, indicating that its a better initialization for RLVR.

    \item \textbf{Methodologically, we find that higher pre-RLVR accuracy does not necessarily translate into higher post-RLVR accuracy.} In our experiments, Mixed SFT records the lowest pre-RLVR accuracy among all training strategies yet ends with the highest post-RLVR ceiling. This highlights the need to evaluate no-CoT training strategies in the context of the full post-training pipeline rather than at intermediate checkpoints.

    \item \textbf{Analytically, we uncover the mechanisms behind different no-CoT training outcomes.} Our analysis suggests that next-chunk reasoning objectives can collapse into local completion, while Mixed SFT preserves no-CoT signal that RLVR can amplify. These findings provide practical guidance for future work on reasoning-rich but trace-free data.
\end{itemize}

%% file: sec/3_experimental_setup.tex
\section{Experimental Setup}
\label{sec:experimental_setup}

\paragraph{Models.}
Qwen3-30B-A3B-Base~\citep{yang2025qwen3} serves as the base model throughout our experiments. We opt for a base checkpoint rather than an instruction-tuned one to ensure a clean, unified initialization free from prior post-training confounds.

\paragraph{Training data.}
All training data is crawled from AoPS~\citep{aops}. A subset of the collected problems is annotated with DeepSeek-V3.2~\citep{liu2025deepseek} to construct \emph{long-CoT data}: we generate reasoning trajectories and retain only those whose final answer is correct, yielding 152K trajectories ($\approx$1.95B tokens). Problems with brief original AoPS solutions, containing derivations but no explicit reasoning traces, serve as \emph{no-CoT data}, totaling 421K solutions ($\approx$0.53B tokens). A representative long-CoT trajectory and no-CoT solution are shown in Fig.~\ref{fig:case_longcot} and Fig.~\ref{fig:case_nocot}. For the subsequent RLVR stage, we use DAPO-Math-17K~\citep{yu2026dapo} as the RL training set.

\paragraph{Training strategies.}
We compare five strategies: NCR at two granularities---NTR, instantiated with RPT~\citep{dong2025reinforcement}, and NSR, instantiated with RLPT~\citep{li2025reinforcement}---along with Sequential SFT, Mixed SFT, and Reasoning SFT as a long-CoT-only baseline. Both NTR and NSR are initialized from Reasoning SFT. We exclude a standalone no-CoT SFT stage layered on top of Reasoning SFT, as this disrupts the long-CoT format and yields format-incoherent outputs. All five strategies are followed by the same RLVR stage using Group Relative Policy Optimization (GRPO)~\citep{shao2024deepseekmath} with rule-based exact-match rewards on DAPO-Math-17K~\citep{yu2026dapo}; full hyperparameter details are in the appendix.

\paragraph{Evaluation.}
We evaluate each model both before and after the RLVR stage, treating post-RLVR accuracy as the primary metric, as it reflects the ceiling each initialization enables, across in-domain (ID) and out-of-domain (OOD) benchmarks. The ID benchmarks are six competition-mathematics sets: AIME 2024/2025/2026~\citep{zhang2024aime24, aime25, aime26}, HMMT 2025/2026~\citep{dekoninck2026beyond}, and IMO-Answer~\citep{luong2025towards}. The OOD benchmarks are HLE~\citep{phan2025humanity}, GPQA-Diamond~\citep{rein2023gpqa}, and MMLU-Pro~\citep{wang2024mmlu}. We report avg@32 on AIME, HMMT, and IMO-Answer, avg@4 on GPQA-Diamond, and pass@1 on HLE and MMLU-Pro.

\begin{table*}[t]
\centering
\scriptsize
\setlength{\tabcolsep}{3.5pt}
\caption{
Performance of the compared no-CoT data utilization strategies, before and after RLVR, on in-domain (ID) and out-of-domain (OOD) reasoning benchmarks.
}
\label{tab:main_results}
\resizebox{\textwidth}{!}{
\begin{tabular}{lcccccc|ccc}
\toprule
\multirow{2}{*}{Method} & \multicolumn{6}{c}{ID Reasoning} & \multicolumn{3}{c}{OOD Reasoning} \\
\cmidrule(lr){2-7} \cmidrule(lr){8-10}
& AIME24 & AIME25 & AIME26 & HMMT25 & HMMT26 & IMO-Ans. & HLE & GPQA-Dia. & MMLU-Pro \\
\midrule
Reasoning SFT
& 73.33 & 64.17 & 54.42 & 34.16 & 32.88 & 34.50 & 6.91 & 46.21 & 74.89 \\
\quad + RLVR
& 81.98 & 76.67 & 65.10 & 43.33 & 39.39 & 47.25 & 7.22 & 56.94 & 73.95 \\
\midrule
Sequential SFT
& 71.25 & 62.81 & 54.42 & 34.17 & 32.88 & 35.75 & 6.52 & 46.21 & 68.82 \\
\quad + RLVR
& 74.90 & 69.27 & 56.35 & 32.92 & 35.23 & 41.25 & 6.79 & 54.55 & 71.10 \\
\midrule
Mixed SFT
& 45.52 & 42.50 & 20.94 & 10.42 & 15.91 & 12.50 & 6.71 & 36.24 & 56.61 \\
\quad + RLVR
& \textbf{87.50} & \textbf{85.73} & 70.42 & \textbf{55.00} & \textbf{51.52} & \textbf{54.00} & \textbf{9.24} & \textbf{60.98} & \textbf{75.84} \\
\midrule
Reasoning SFT + NTR
& 74.69 & 67.29 & 53.75 & 36.77 & 34.38 & 38.00 & 7.59 & 53.66 & 73.46 \\
\quad + RLVR
& \textbf{87.50} & 84.38 & 69.38 & 50.42 & 47.16 & 46.50 & 7.68 & 57.70 & 74.03 \\
\midrule
Reasoning SFT + NSR
& 74.27 & 63.54 & 56.88 & 38.33 & 38.33 & 36.00 & 7.46 & 42.93 & 70.31 \\
\quad + RLVR
& 85.52 & 80.92 & \textbf{72.19} & 48.85 & 46.02 & 48.25 & 7.72 & 56.31 & 74.89 \\
\bottomrule
\end{tabular}
}
\end{table*}

%% file: sec/4_main_results.tex
\section{Main Results}
\label{sec:main_results}

Tab.~\ref{tab:main_results} summarizes the post-RLVR performance of all five training strategies. We draw three key findings from this comparison and examine each in the following subsections:
\begin{enumerate}
    \renewcommand{\labelenumi}{(\arabic{enumi})}
    \item \textbf{Mixed SFT is a stronger alternative to next-chunk reasoning RL.} It achieves better post-RLVR performance on both in-domain and out-of-domain benchmarks.
    \item \textbf{Mixed SFT clearly outperforms Sequential SFT}, indicating that the way no-CoT data is incorporated matters.
    \item \textbf{Pre-RLVR accuracy is not a reliable indicator of post-RLVR potential.} A model with higher intermediate accuracy does not necessarily reach a higher final performance after RLVR.
\end{enumerate}

\subsection{Paradigm: SFT vs.\ next-chunk reasoning}

We compare SFT and next-chunk reasoning RL along three axes: in-domain accuracy, OOD generalization, and training efficiency. Mixed SFT outperforms next-chunk reasoning RL on all three.

\begin{figure}[t]
\centering
\includegraphics[width=1\columnwidth]{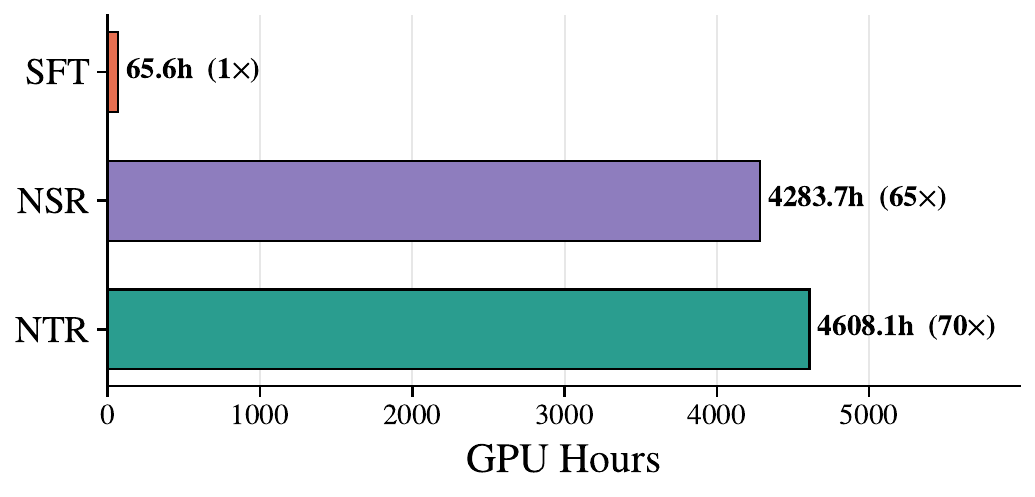}
\caption{GPU-hour training cost of each strategy on the same no-CoT data. NTR and NSR are measured at 160 steps, where both reach peak pre-RLVR accuracy.}
\label{fig:cost}
\end{figure}

\paragraph{In-domain reasoning.}
As summarized in Tab.~\ref{tab:main_results}, Mixed SFT averages 67.4 post-RLVR across the six in-domain benchmarks (AIME 24/25/26, HMMT 25/26, IMO-Answer), 3.1 points above the next-best NTR (64.2) and 3.7 points above NSR (63.6). In short, SFT can effectively leverage no-CoT data without an explicit RL-based reasoning-reconstruction objective.

\paragraph{OOD generalization.}
On the three OOD benchmarks HLE~\citep{phan2025humanity}, GPQA-Diamond~\citep{rein2023gpqa}, and MMLU-Pro~\citep{wang2024mmlu}, Mixed SFT reaches the highest post-RLVR accuracy of 9.24, 60.98, and 75.84 respectively, surpassing both NTR and NSR by clear margins on each. Mixed SFT therefore outperforms next-chunk reasoning RL on OOD benchmarks as well, indicating that Mixed SFT's advantage is not a math-specific overfit but extends to broader reasoning settings.

\begin{figure}[t]
    \centering
    \includegraphics[width=\columnwidth]{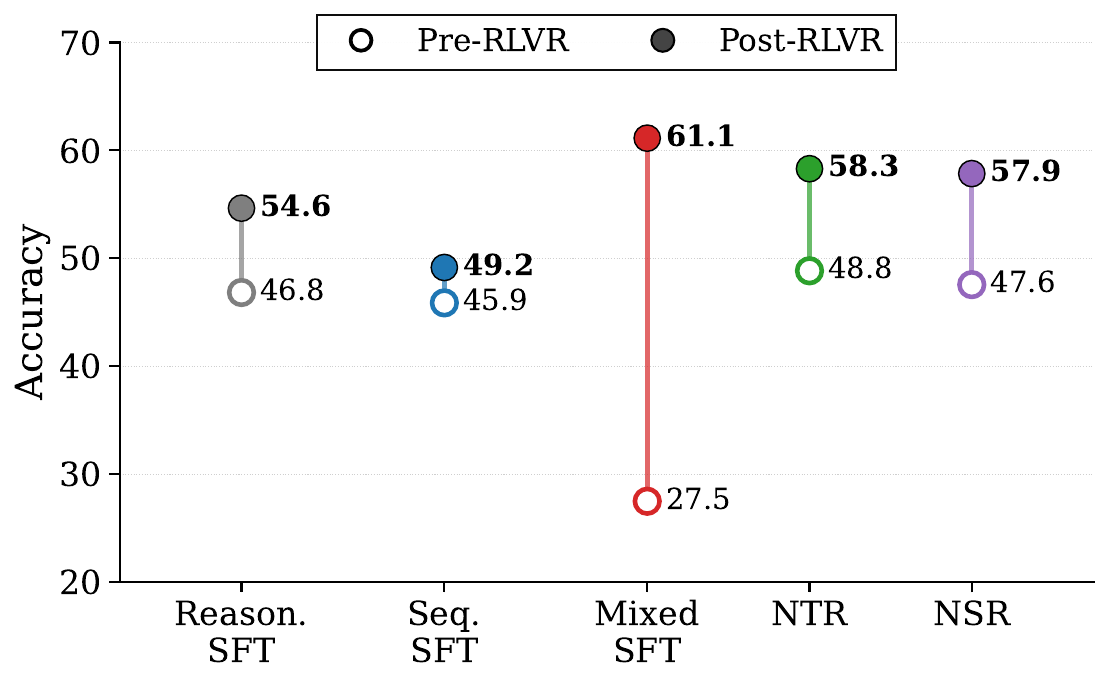}
    \caption{
    Pre-RLVR and post-RLVR accuracy averaged over all nine reasoning benchmarks, shown as per-method dumbbells.
    }
    \label{fig:pre_vs_post}
\end{figure}

\begin{figure*}[t]
    \centering
    \includegraphics[width=1\textwidth]{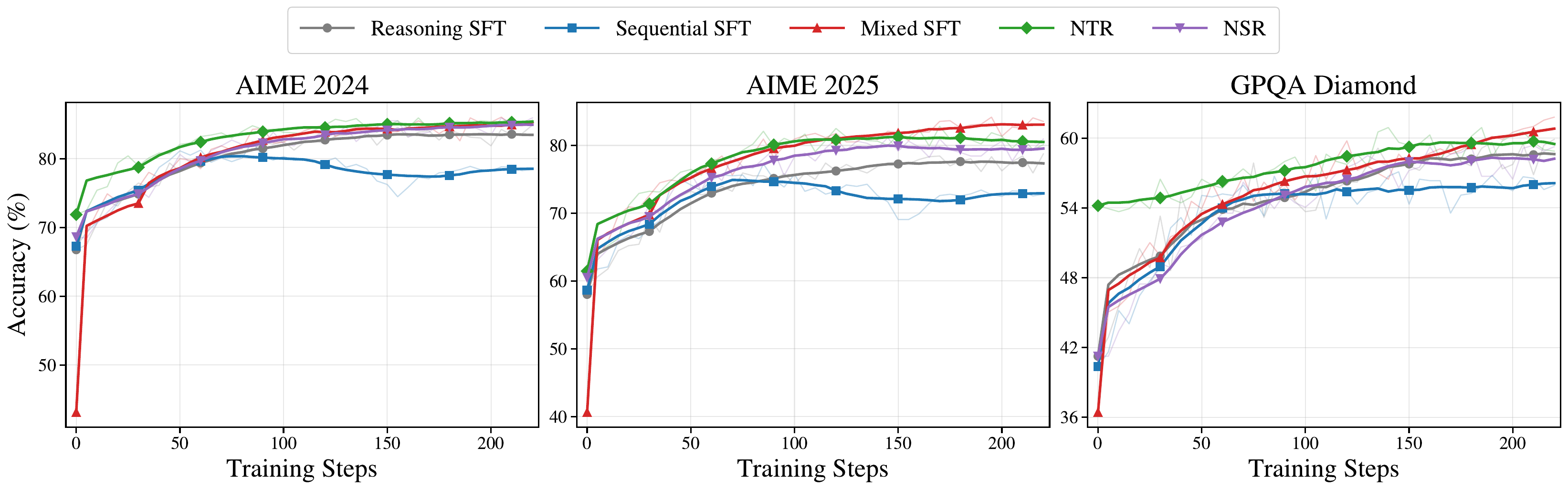}
    \caption{
    Post-RLVR training curves of all five strategies on AIME 2024, AIME 2025, and GPQA-Diamond under the same RLVR stage.
    }
    \label{fig:rlvr_curves}
\end{figure*}

\paragraph{Training efficiency.}
SFT is also substantially more efficient. As shown in Fig.~\ref{fig:cost}, both NTR and NSR consume over 60$\times$ more GPU hours than SFT on the same no-CoT data, since both require online rollouts, reward computation, and policy optimization, with NSR further depending on a generative reward model. The efficiency gap thus compounds the performance gap: next-chunk reasoning RL neither outperforms SFT nor is cheaper than it for leveraging no-CoT data.

\subsection{Data combination: Mixed vs.\ Sequential SFT}

Within SFT, Mixed SFT achieves a significantly higher final ceiling than Sequential SFT, showing that for no-CoT data, how it is organized with reasoning data matters as much as whether it is included. As shown in Fig.~\ref{fig:rlvr_curves}, Mixed SFT starts from the lowest pre-RLVR accuracy and trails the other methods early in RLVR training, but gradually catches up and overtakes them, ending with the highest ceiling on all three representative benchmarks.

Sequential SFT often records better pre-RLVR accuracy because its final stage is Reasoning SFT, whose long-CoT format aligns with evaluation; however, this stage also partially overwrites what was learned from no-CoT data, a phenomenon we quantify in Observation~6 of Sec.~\ref{sec:analysis} via a post-RLVR retention probe on no-CoT problems. Mixed SFT instead exposes the model to both data types simultaneously: pre-RLVR accuracy drops due to the conflicting output structures, but no cross-stage forgetting occurs, preserving mathematical knowledge, derivation paths, and problem variants. RLVR then optimizes answer correctness from this richer initialization, yielding a higher final ceiling.

\subsection{Evaluation timing: pre- vs.\ post-RLVR scores}

Pre-RLVR scores do not reliably predict post-RLVR performance. As Fig.~\ref{fig:pre_vs_post} shows, Mixed SFT has by far the lowest pre-RLVR accuracy of 27.5---roughly 20 points below every other method---yet the highest post-RLVR accuracy of 61.1, a pre-to-post improvement of 33.7 points that is more than three times larger than any other method; the remaining four methods start from similar pre-RLVR levels between 45.9 and 48.8 and rise by 3.3 to 10.3 points. No-CoT data may not immediately improve explicit reasoning outputs, but it reshapes the model's internal knowledge distribution, and RLVR amplifies these latent capabilities through verifiable rewards. The right criterion for evaluating a no-CoT initialization strategy is therefore not the immediate pre-RLVR score, but the post-RLVR ceiling it enables.

%% file: sec/5_analysis.tex
\section{Analysis}
\label{sec:analysis}

\begin{figure}[t]
    \centering
    \includegraphics[width=0.98\columnwidth]{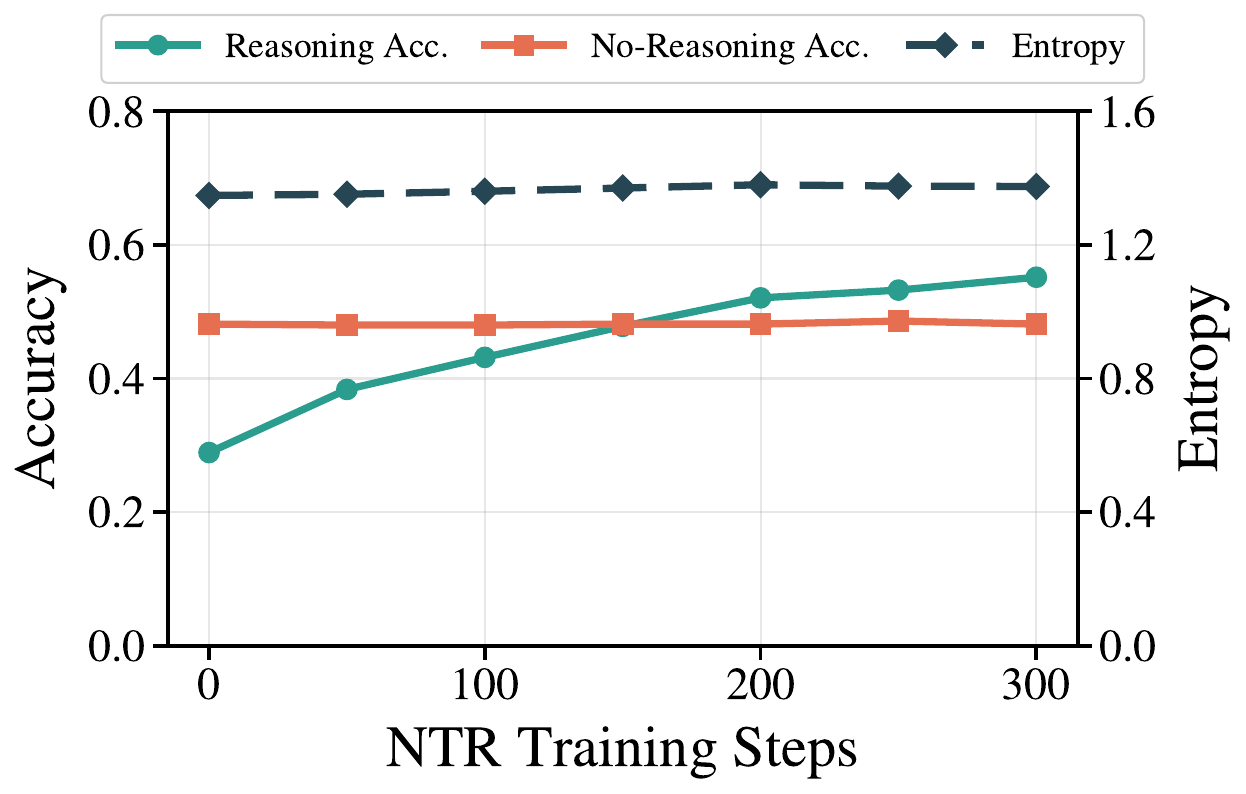}
    \caption{
    Per-token entropy, reasoning accuracy, and no-reasoning accuracy on $2{,}048$ randomly sampled high-entropy tokens, measured with NTR checkpoints across training steps.
    }
    \label{fig:highentropy_dynamics}
\end{figure}

\begin{figure*}[t]
\centering
\begin{subfigure}[b]{0.472\textwidth}
    \centering
    \includegraphics[width=\textwidth]{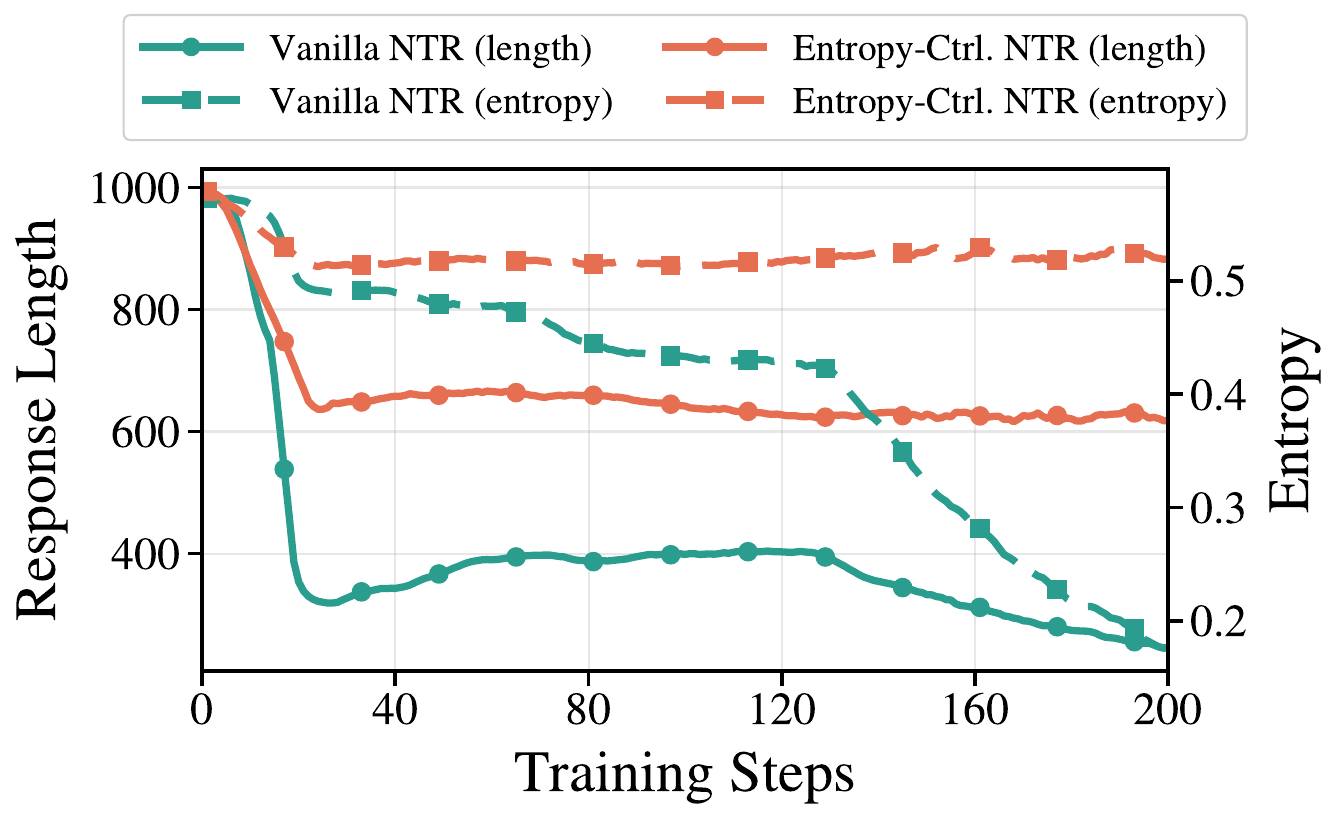}
    \caption{Training dynamics across steps.}
    \label{fig:rpt_entropy}
\end{subfigure}
\hfill
\begin{subfigure}[b]{0.508\textwidth}
    \centering
    \includegraphics[width=\textwidth]{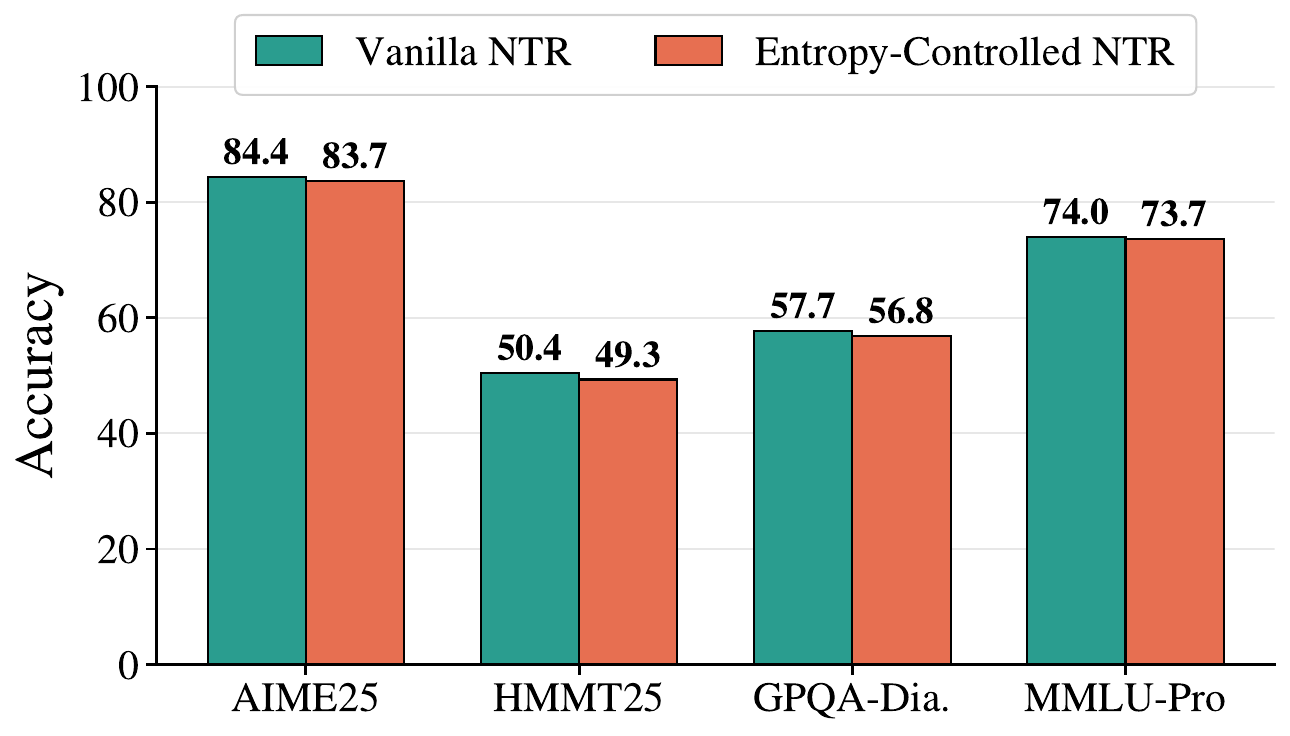}
    \caption{Post-RLVR accuracy.}
    \label{fig:rpt_entropy_control}
\end{subfigure}
\caption{
\textbf{(\subref{fig:rpt_entropy})} Generation entropy and response length of vanilla NTR and the entropy-controlled variant across training steps.
\textbf{(\subref{fig:rpt_entropy_control})} Post-RLVR accuracy of the two variants on four representative benchmarks; per-benchmark numbers are in Tab.~\ref{tab:rpt_entropy_control}.
}
\label{fig:rpt_dynamics}
\end{figure*}

This section addresses three questions raised by the main results:
\begin{itemize}
    \item First, does next-chunk reasoning RL bring genuine reasoning gains beyond what SFT on the same no-CoT data already achieves?
    \item Second, why does Mixed SFT recover from the lowest pre-RLVR accuracy to the highest post-RLVR ceiling?
    \item Third, why does Sequential SFT, which sees the same no-CoT and long-CoT data, end up well below Mixed SFT?
\end{itemize}
For the first question, we examine NTR from three angles---the targets it selects for supervision, the reasoning traces it produces, and whether its ceiling can be lifted by preserving exploration---and then rule out a token-coverage confound. For the second, we trace Mixed SFT's pre-RLVR drop to a transient format mismatch that RLVR repairs. For the third, we use a post-RLVR retention probe to show that Sequential SFT's second long-CoT stage partially erases what its first no-CoT stage absorbed, and that RLVR cannot recover the lost signal.

\noindent\textbf{Observation 1: The Entropy Filter Does Not Select Reasoning-Hard Tokens.}
NTR rests on the claim that high-entropy tokens are the hard, reasoning-demanding targets worth optimizing, and accordingly supervises only the top-20\% highest-entropy tokens of the no-CoT corpus~\citep{dong2025reinforcement}. We test this claim directly. We randomly draw $2{,}048$ tokens from the high-entropy pool and, across NTR checkpoints saved at increasing training steps, measure three quantities at these positions: the per-token entropy, the prediction accuracy with an explicit reasoning step (\emph{reasoning accuracy}), and the accuracy of direct prediction (\emph{no-reasoning accuracy}). As Fig.~\ref{fig:highentropy_dynamics} shows, reasoning accuracy climbs steadily from $0.29$ to $0.55$ while the entropy of the same tokens stays almost flat; if entropy tracked predictive difficulty, such a large accuracy gain would be mirrored by a comparable entropy change. The no-reasoning accuracy is more direct evidence: it stays around $0.48$ throughout, so nearly half of these tokens are already predicted correctly without any reasoning. High token entropy therefore reflects the local uncertainty of the raw corpus, not reasoning difficulty, and the entropy filter still admits a large fraction of tokens that demand no reasoning at all.

\noindent\textbf{Observation 2: NTR's Reasoning Traces Degenerate into Local Completion.}
Because many of the filtered targets are in fact locally predictable, NTR can satisfy its reconstruction reward without genuine reasoning. Fig.~\ref{fig:rpt_entropy} shows the consequence: both generation entropy and response length decrease steadily during NTR training, indicating that the model's outputs become shorter and more deterministic as training proceeds. Inspecting the generated traces confirms this---the model converges to a few repetitive templates across very different prefixes (Fig.~\ref{fig:rpt_case}, Appendix), performing a brief, template-like local completion of the immediate context rather than long-horizon reasoning. NSR exhibits the same tendency at the sentence level, where the predicted sentence is often a direct continuation of the preceding text; a representative NSR trajectory is shown in Fig.~\ref{fig:case_rlpt} (Appendix). In both cases the reconstruction reward is met by local pattern completion, not by the reasoning the objective is meant to elicit. This template collapse leaves NTR with little reasoning structure for the downstream RLVR stage to amplify.

\noindent\textbf{Observation 3: Preserving Entropy Does Not Raise NTR's Ceiling.}
The entropy collapse observed in NTR suggests a natural rescue hypothesis: perhaps NTR's mediocre post-RLVR ceiling is caused by the collapse, and preserving exploration would let the model learn richer reasoning and reach a higher ceiling. We test this by modifying NTR training with two interventions that suppress the collapse, both detailed in App.~\ref{appendix:entropy_control}: rollout groups with high in-group success rate are stochastically dropped so that already-easy prompts cannot push the policy toward over-confident outputs~\citep{luo2026compress, xiong2025reinforce}, and positive advantages are down-weighted relative to negative ones to keep the policy from converging onto a single template~\citep{yu2026dapo, wang2025aspo}. As Fig.~\ref{fig:rpt_entropy} shows, the interventions work: generation entropy stays near $0.52$ and response length around $640$ tokens throughout training. Yet Fig.~\ref{fig:rpt_entropy_control} shows that this entropy-controlled NTR reaches a slightly \emph{lower} post-RLVR ceiling than vanilla NTR on every benchmark. Entropy collapse is therefore not the cause of NTR's mediocre ceiling; vanilla NTR reaches its ceiling \emph{by} converging onto the peaked, template-like policy, and preventing that convergence only removes a usable solution without supplying a better one.

\begin{table}[t]
\centering
\scriptsize
\setlength{\tabcolsep}{3.5pt}
\caption{
Post-RLVR performance when NTR or NSR is inserted between Mixed SFT and RLVR, compared with Mixed SFT directly followed by RLVR.
}
\label{tab:mixed_sft_reconstruction}
\resizebox{\columnwidth}{!}{
\begin{tabular}{lcccc}
\toprule
Method & AIME24 & AIME25 & HMMT25 & HMMT26 \\
\midrule
Mixed SFT
& 45.52 & 42.50 & 10.42 & 15.91 \\
+ RLVR
& \textbf{87.50} & \textbf{85.73} & \textbf{55.00} & \textbf{51.52} \\
+ NTR + RLVR
& 86.35 & 84.58 & 53.73 & 50.50 \\
+ NSR + RLVR
& 87.21 & 84.67 & 53.15 & 49.52 \\
\bottomrule
\end{tabular}
}
\end{table}

\vspace{1em}
\noindent\textbf{Observation 4: NTR and NSR Add No Gain on Top of Mixed SFT.}
A final concern is a confound: next-chunk reasoning RL is far more expensive than SFT and therefore covers far fewer tokens of the no-CoT corpus, so SFT's advantage might come from larger token exposure rather than from the training paradigm. To rule this out, we continue NTR or NSR training from the Mixed SFT model, which has already absorbed the full no-CoT corpus, and then apply the same RLVR stage. As Tab.~\ref{tab:mixed_sft_reconstruction} shows, inserting an NTR or NSR stage between Mixed SFT and RLVR leaves final performance almost unchanged, and slightly lower on several tasks. Once the model has been sufficiently exposed to no-CoT data through SFT, neither objective adds reasoning ability that RLVR can amplify. Taken together, these analyses suggest that the improvements of next-chunk reasoning RL over Reasoning SFT are not primarily delivered by its reasoning-reconstruction mechanism. A substantial fraction of its supervised targets are locally predictable, the generated traces tend toward template-like local completion, preserving exploration does not raise the ceiling, and once the same no-CoT corpus has been absorbed through SFT, an additional next-chunk reasoning RL stage no longer adds gains that RLVR can amplify.

\begin{figure}[t]
    \centering
    \includegraphics[width=0.98\columnwidth]{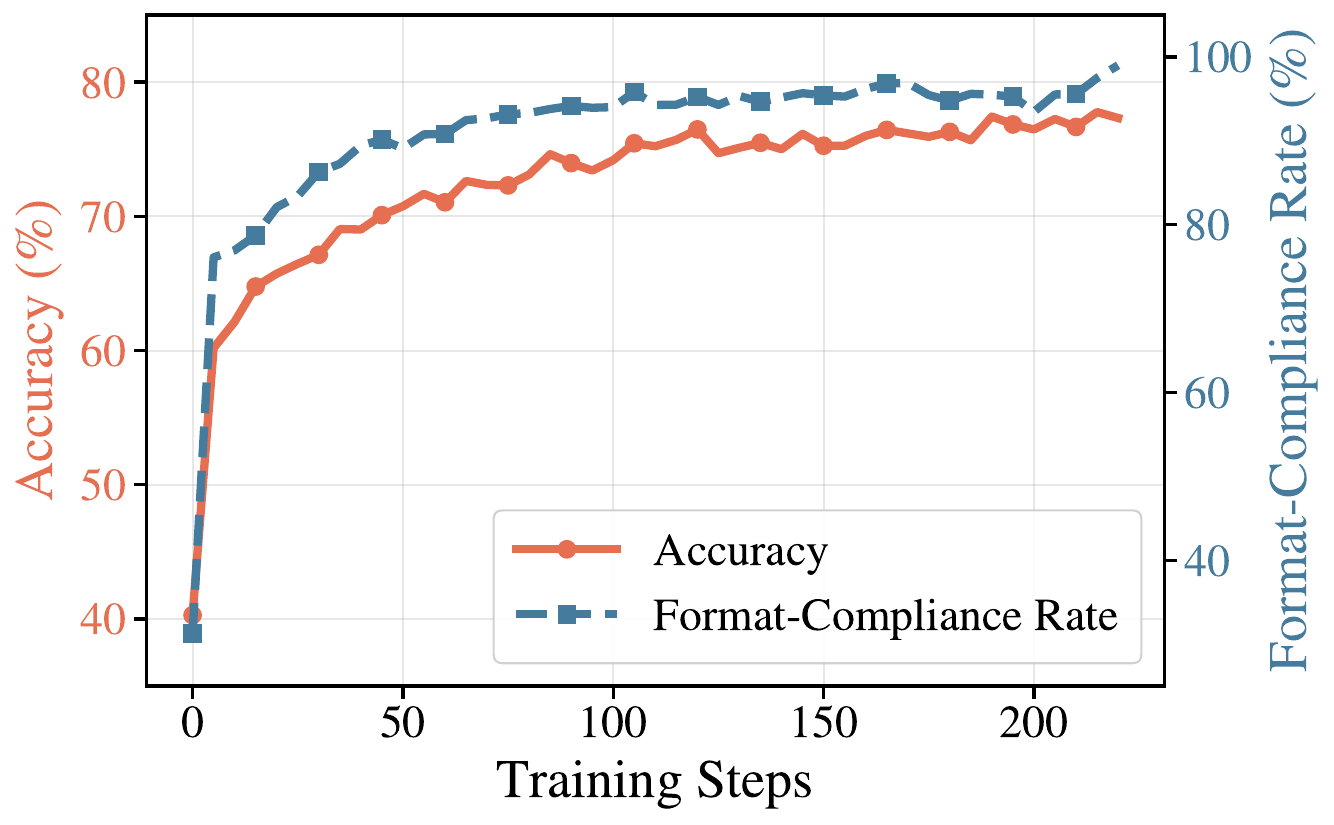}
    \caption{
    Accuracy and format-compliance rate of Mixed SFT across RLVR training steps.
    }
    \label{fig:accuracy_format_ratio}
\end{figure}

\vspace{1em}
\noindent\textbf{Observation 5: Mixed SFT's Pre-RLVR Drop Is a Transient Format Artifact.}
We now turn to the second question: why Mixed SFT, despite the lowest pre-RLVR accuracy, reaches the highest post-RLVR ceiling. The pre-RLVR drop comes from an output-structure mismatch between the two data types. Long-CoT data carries explicit thinking markers, long reasoning traces, and standardized answer formats, whereas no-CoT data resembles ordinary mathematical solutions with no unified reasoning boundaries or answer format. Training on both jointly temporarily destabilizes the output structure: the explicit reasoning process is sometimes skipped entirely, and the response format becomes inconsistent (Fig.~\ref{fig:case_mixed_sft}, Appendix). This depresses pre-RLVR accuracy but does not erase what the model has learned. As Fig.~\ref{fig:accuracy_format_ratio} shows, during RLVR the format-compliance rate rises quickly and accuracy grows alongside it: verifiable rewards re-impose a consistent output format, while the model draws on the broader mathematical knowledge, derivation patterns, and problem variants absorbed during Mixed SFT, ultimately reaching a higher ceiling. The pre-RLVR drop is thus a transient format artifact, not a loss of capability, which is why pre-RLVR performance underestimates Mixed SFT and cannot predict the post-RLVR upper bound.

\begin{table}[t]
\centering
\small
\setlength{\tabcolsep}{6pt}
\caption{
Post-RLVR avg@8 on No-CoT Probe, a 100-problem retention probe randomly drawn from the no-CoT training corpus.
}
\label{tab:no_cot_retention}
\begin{tabular}{lc}
\toprule
Method & No-CoT Probe \\
\midrule
Sequential SFT $+$ RLVR & 59.19 \\
Mixed SFT $+$ RLVR      & \textbf{68.63} \\
\bottomrule
\end{tabular}
\end{table}

\vspace{1em}
\noindent\textbf{Observation 6: Sequential SFT Forgets No-CoT Knowledge Across Stages.}
The previous subsection asked why Mixed SFT recovers from its pre-RLVR drop. We now ask why Sequential SFT, which sees the same no-CoT and long-CoT data, ends up below Mixed SFT after RLVR. A direct pre-RLVR comparison would be unreliable, since Mixed SFT's pre-RLVR outputs are format-unstable (Fig.~\ref{fig:case_mixed_sft}, Appendix) while Sequential SFT's are not; we therefore evaluate after the same RLVR stage, which equalizes the output format (Fig.~\ref{fig:accuracy_format_ratio}). We randomly sample 100 problems from the no-CoT training corpus as a retention probe---both pipelines saw these problems during their SFT stages, so any post-RLVR gap reflects how each preserved the no-CoT signal across SFT. We sample 8 responses per problem and report avg@8 in Tab.~\ref{tab:no_cot_retention}. Mixed SFT outperforms Sequential SFT on the probe, consistent with the view that Sequential SFT's second long-CoT stage overwrites part of the no-CoT signal absorbed in its first stage, and that RLVR does not recover knowledge erased before it began.

%% file: sec/2_related.tex
\section{Related Work}

\paragraph{SFT for reasoning post-training.}
Supervised fine-tuning is the dominant way to inject data into post-training. Early SFT relied on instruction-response pairs (FLAN~\citep{wei2021finetuned}, LIMA~\citep{zhou2023lima}); with the rise of reasoning models, SFT was extended to explicit CoT data through bootstrapping (STaR~\citep{zelikman2022star}), distillation (Distilling Step-by-Step~\citep{hsieh2023distilling}), and large-teacher transfer (DeepSeek-R1-Distill~\citep{guo2025deepseek}). Several recent works further show that mixing long-CoT and non-CoT data within a single SFT stage is itself effective: DeepSeek-R1's second-stage SFT recovers non-reasoning ability~\citep{guo2025deepseek}; Qwen3's Thinking Mode Fusion~\citep{yang2025qwen3} jointly trains long-CoT and short responses; and LS-Mixture~\citep{yu2025long} pairs long-CoT trajectories with structure-preserved short rewrites to mitigate overthinking. These works, however, all place mixed SFT \emph{after} reasoning-oriented RL, treating the mixture as a recovery step or a stand-alone post-training recipe. Our work studies the complementary question: whether Mixed SFT on long-CoT plus raw no-CoT text is the right way to initialize a subsequent RLVR stage, by systematically comparing Sequential SFT, Mixed SFT, and their combination with subsequent RLVR.

\paragraph{Next-chunk reasoning RL.}
Reinforcement learning has become a central paradigm for improving reasoning. RLHF~\citep{christiano2017deep} aligns models with human preferences, while RLVR~\citep{guo2025deepseek, team2025kimi, shao2024deepseekmath} uses rule-based rewards over verifiable answers and has substantially advanced math and coding. Both depend on specialized data, namely costly preference annotations or verifiable-answer corpora, which motivates attempts to extend RL to broader natural text. Next-chunk reasoning RL converts no-CoT text into an RL-optimizable task: the model first generates reasoning, then is rewarded for predicting subsequent content. RPT~\citep{dong2025reinforcement} reformulates next-token prediction as next-token reasoning with prefix-matching rewards, but compares only against next-token-prediction continuation on the same no-CoT data and starts from a post-trained model with long-CoT behavior already built in. RLP~\citep{hatamizadeh2025rlp} moves this token-level objective into pretraining. RLPT~\citep{li2025reinforcement} raises the granularity from tokens to sentences with a generative reward model, but reports no SFT-only comparison. RMT~\citep{tian2025reinforcement} brings the idea into mid-training with dynamic token budgets and curriculum sampling, also without combining no-CoT with long-CoT in its SFT baseline. PretrainZero~\citep{xing2025pretrainzero} extends this idea to span-level prediction starting from a base model, with an SFT baseline restricted to single-source no-CoT data. As Tab.~\ref{tab:cot_recon_comparison} summarizes, none of these works evaluates next-chunk reasoning RL against an SFT baseline that combines no-CoT with long-CoT data, exactly the gap our work fills.

%% file: sec/6_conclusion.tex
\section{Discussion and Conclusion}

This paper revisits whether next-chunk reasoning RL outperforms SFT for utilizing no-CoT data, comparing NTR, NSR, Sequential SFT, Mixed SFT, and Reasoning SFT under a unified RLVR budget from the same pre-trained base across six in-domain and three out-of-domain reasoning benchmarks. Despite the lowest pre-RLVR accuracy, Mixed SFT reaches the highest post-RLVR ceiling on both in-domain and out-of-domain benchmarks while requiring over 60$\times$ less training compute. Mechanistically, next-chunk reasoning RL's gains do not arise from genuine reasoning---its high-entropy targets are largely locally predictable and its traces collapse into template-like completion---while Mixed SFT avoids the cross-stage forgetting that limits Sequential SFT and preserves the no-CoT signal that RLVR later amplifies. Our results also expose a methodological caveat: higher pre-RLVR accuracy does not imply higher post-RLVR accuracy, so no-CoT strategies should be evaluated through the full RLVR pipeline. For practitioners working under compute constraints, this points to data composition within the SFT stage rather than additional RL stages.

\newpage

%% file: sec/7_appendix.tex
\clearpage

\section{Limitations}
\label{sec:limitations}

\paragraph{Mixing ratio in Mixed SFT.}
Our Mixed SFT experiments use a single fixed mixing ratio between no-CoT and long-CoT data. We do not explore how varying this ratio affects the resulting post-RLVR ceiling, the magnitude of the pre-RLVR drop, or the trade-off between absorbing no-CoT knowledge and preserving the long-CoT format. A systematic sweep over mixing ratios---including the regime where one source is scarce---is an important direction for future work and may further refine the recipe-level guidance offered by our analysis.

\paragraph{Domain coverage.}
Both the no-CoT training corpus and all evaluation benchmarks are mathematical. We do not investigate whether the same conclusions transfer to code generation, scientific reasoning outside of GPQA, multilingual tasks, or other domains in which no-CoT data may have different structural and token-level properties. Whether Mixed SFT remains the most effective way to leverage no-CoT data in such settings is an open question.

\section{Training Details}
\label{appendix:training}

\paragraph{Framework and hardware.}
All SFT and RL experiments are run on NVIDIA H200 GPUs. Every RL stage (NTR, NSR, and RLVR) uses 64 GPUs.

\paragraph{Training data.}
The long-CoT corpus contains 152K examples and roughly 1.95B tokens: queries are drawn from AoPS~\citep{aops}, reasoning trajectories are produced by DeepSeek-V3.2~\citep{liu2025deepseek} with the maximum generation length set to 64K tokens, and only trajectories with correct final answers are retained. The no-CoT corpus contains 421K examples and roughly 0.53B tokens, also crawled from AoPS. For the subsequent RLVR stage we use DAPO-Math-17K~\citep{yu2026dapo} as the RL training set. We additionally deduplicate all training data (long-CoT, no-CoT, and DAPO-Math-17K) against the evaluation benchmarks listed in App.~\ref{appendix:evaluation} to prevent contamination.

\paragraph{SFT hyperparameters.}
All SFT stages (Reasoning SFT, Sequential SFT, and Mixed SFT) are trained for a single epoch with a maximum sequence length of $131{,}072$ tokens.

\paragraph{RL hyperparameters.}
All RL stages share the same optimizer configuration: learning rate $1\text{e}{-}6$, and the KL penalty against the reference model is disabled. The remaining stage-specific hyperparameters are listed in Tab.~\ref{tab:rl_hyperparams}.

\begin{table}[t]
\centering
\small
\setlength{\tabcolsep}{5pt}
\caption{
RL hyperparameters for NTR, NSR, and RLVR. Max response length is reported in tokens.
}
\label{tab:rl_hyperparams}
\begin{tabular}{lccc}
\toprule
Setting & NTR & NSR & RLVR \\
\midrule
Global batch size  & 4096 & 256 & 256 \\
Rollouts per prompt         & 64   & 16  & 16  \\
Max response length         & 2K   & 32K & 32K \\
Learning rate               & $1\text{e}{-}6$ & $1\text{e}{-}6$ & $1\text{e}{-}6$ \\
KL loss                     & off  & off & off \\
\bottomrule
\end{tabular}
\end{table}

\paragraph{Reward models.}
NTR uses token-level prefix-matching against the ground-truth continuation of the no-CoT corpus as the reward signal. NSR uses gpt-oss-120b~\citep{agarwal2025gpt} with \texttt{reasoning\_effort=high} as a generative judge for sentence-level semantic consistency between the model's prediction and the ground-truth segment. RLVR uses rule-based exact-match rewards against the verifiable answers in DAPO-Math-17K.

\paragraph{Reconstruction targets.}
NTR and NSR differ in how they sample reconstruction targets from the no-CoT corpus. For NTR, we first run Qwen3-30B-A3B over the full no-CoT corpus to obtain per-token next-token entropies, and only keep the top-20\% highest-entropy tokens as NTR prediction targets, on the assumption that low-entropy tokens are too predictable from local context to provide a meaningful learning signal. For NSR, we segment each no-CoT example into sentences and use the full set of segmented sentences as reconstruction targets, without entropy-based filtering.

\paragraph{Training objectives.}
We summarize here the per-step objectives of the SFT, NTR, and NSR stages compared in this work; all three are followed by the same RLVR stage. Let $\pi_\theta$ denote the policy parameterized by $\theta$.

\textit{SFT.} Given a dataset $\mathcal{D}_{\text{SFT}}$ of (prompt, response) pairs $(x, y)$, the SFT loss is the standard token-level negative log-likelihood,
\begin{multline}
\mathcal{L}_{\text{SFT}}(\theta) = \\
-\mathbb{E}_{(x, y) \sim \mathcal{D}_{\text{SFT}}}\!\left[\frac{1}{|y|}\sum_{t=1}^{|y|}\log \pi_\theta(y_t \mid x, y_{<t})\right].
\end{multline}
The three SFT strategies in our comparison differ only in $\mathcal{D}_{\text{SFT}}$: Reasoning SFT uses long-CoT data only, Sequential SFT runs two consecutive SFT stages (one on no-CoT, one on long-CoT), and Mixed SFT uses the union of both in a single stage.

\textit{NTR.} Following~\citet{dong2025reinforcement}, NTR reframes next-token prediction as a reasoning task. Let $\mathcal{D}_{\text{NTR}}$ be the entropy-filtered prefix set defined above. For each prefix $x_{<t} \in \mathcal{D}_{\text{NTR}}$, the policy generates $G$ rollouts $o^i = (c^i, y^i) \sim \pi_\theta(\cdot \mid x_{<t})$, where $c^i$ is the reasoning trace and $y^i$ is the predicted continuation. With prefix-matching reward against the ground-truth continuation $x_{\geq t}$,
\begin{equation}
r^i =
\begin{cases}
1, & \bar{y}^i = \bar{x}_{\geq t}[1{:}l]\;\text{and}\;l \in \mathcal{L}_{\text{gt}}, \\
0, & \text{otherwise},
\end{cases}
\end{equation}
where $\bar{\cdot}$ denotes the byte sequence of a token sequence, $l = |\bar{y}^i|$, and $\mathcal{L}_{\text{gt}}$ is the set of cumulative byte lengths of tokens in $x_{\geq t}$. The NTR objective is the expected prefix-matching reward,
\begin{equation}
\mathcal{J}_{\text{NTR}}(\theta) = \mathbb{E}_{x_{<t} \sim \mathcal{D}_{\text{NTR}},\; \{o^i\} \sim \pi_\theta(\cdot \mid x_{<t})}\!\big[\,r^i\,\big].
\end{equation}

\textit{NSR.} Following~\citet{li2025reinforcement}, NSR replaces token-level prediction with sentence-level prediction. Let $\mathcal{D}_{\text{NSR}} = \{(s_{<i}, s_i)\}$ be the sentence-segmented no-CoT corpus, where $s_{<i} = [s_1, \ldots, s_{i-1}]$ is the preceding context and $s_i$ is the target sentence. For each $(s_{<i}, s_i)$, the policy generates a rollout $o = (c, \hat{s}_i) \sim \pi_\theta(\cdot \mid s_{<i})$, with $c$ the reasoning trace and $\hat{s}_i$ the predicted next sentence extracted from the response. The reward is the binary judgment of a generative reward model $G_{\text{rm}}$,
\begin{equation}
r = G_{\text{rm}}(\hat{s}_i, s_i) \in \{0, 1\},
\end{equation}
and the NSR objective is
\begin{equation}
\mathcal{J}_{\text{NSR}}(\theta) = \mathbb{E}_{(s_{<i},s_i) \sim \mathcal{D}_{\text{NSR}},\; o \sim \pi_\theta(\cdot \mid s_{<i})}\!\big[\,r\,\big].
\end{equation}

Both $\mathcal{J}_{\text{NTR}}$ and $\mathcal{J}_{\text{NSR}}$ are optimized via GRPO~\citep{shao2024deepseekmath}, and the same GRPO setup is reused for the subsequent RLVR stage on DAPO-Math-17K with rule-based exact-match rewards.

\paragraph{Licenses and intended use.}
The base model Qwen3-30B-A3B-Base~\citep{yang2025qwen3}, the long-CoT trajectory generator DeepSeek-V3.2~\citep{liu2025deepseek}, the generative reward model gpt-oss-120b~\citep{agarwal2025gpt}, the no-CoT corpus crawled from AoPS~\citep{aops}, the DAPO-Math-17K RL training set~\citep{yu2026dapo}, and the evaluation benchmarks listed in App.~\ref{appendix:evaluation} are released under licenses permitting non-commercial research use, and our use is consistent with their stated terms.

\section{Evaluation Details}
\label{appendix:evaluation}

\paragraph{Decoding configuration.}
All benchmarks share the decoding hyperparameters in Tab.~\ref{tab:decoding}. The parsed model answer is extracted from \texttt{\textbackslash boxed\{\}} or from the final-answer span following \texttt{</think>}, and is then compared against the canonical answer via exact match.

\begin{table}[t]
\centering
\small
\setlength{\tabcolsep}{6pt}
\caption{Decoding configuration used for evaluation across all methods and benchmarks.}
\label{tab:decoding}
\begin{tabular}{lc}
\toprule
Parameter & Value \\
\midrule
max tokens   & $64 \times 1024$ \\
top-$k$      & 0 \\
top-$p$      & 0.999 \\
temperature  & 0.8 \\
\bottomrule
\end{tabular}
\end{table}

\begin{table*}[t]
\centering
\scriptsize
\setlength{\tabcolsep}{3.5pt}
\caption{
Per-benchmark pass@$n$ comparison of different no-CoT data utilization strategies before and after RLVR. The setup mirrors Tab.~\ref{tab:main_results} but reports pass@$n$ rather than avg@$n$. $^{\dagger}$HLE and MMLU-Pro are reported as pass@1, i.e., a single sample per question, so their values coincide with those in the main table.
}
\label{tab:main_results_pass_at_n}
\resizebox{\textwidth}{!}{
\begin{tabular}{lccccccccc}
\toprule
Method & AIME24 & AIME25 & AIME26 & HMMT25 & HMMT26 & IMO-Ans. & HLE$^{\dagger}$ & GPQA-Dia. & MMLU-Pro$^{\dagger}$ \\
\midrule
Reasoning SFT
& 90.00 & 90.00 & 83.33 & 76.67 & 63.64 & 68.00 & 6.91 & 68.69 & 74.89 \\
\quad + RLVR
& \textbf{93.33} & \textbf{93.33} & 86.67 & 80.00 & 69.70 & 66.00 & 7.22 & \textbf{77.27} & 73.95 \\
\midrule
Sequential SFT
& 90.00 & 90.00 & 86.67 & 76.67 & 66.67 & 58.00 & 6.52 & 67.17 & 68.82 \\
\quad + RLVR
& 90.00 & 90.00 & 86.67 & 70.00 & 66.67 & 72.00 & 6.79 & 75.76 & 71.10 \\
\midrule
Mixed SFT
& 83.33 & 76.67 & 60.00 & 53.33 & 51.52 & 50.00 & 6.71 & 71.72 & 56.61 \\
\quad + RLVR
& \textbf{93.33} & \textbf{93.33} & 70.42 & \textbf{90.00} & \textbf{87.88} & \textbf{76.00} & \textbf{9.24} & \textbf{77.27} & \textbf{75.84} \\
\midrule
Reasoning SFT + NTR
& 90.00 & 96.67 & 83.33 & 76.67 & 81.82 & 72.00 & 7.59 & 74.24 & 73.46 \\
\quad + RLVR
& \textbf{93.33} & \textbf{93.33} & \textbf{90.00} & \textbf{90.00} & \textbf{87.88} & \textbf{76.00} & 7.68 & 75.76 & 74.03 \\
\midrule
Reasoning SFT + NSR
& 93.33 & 90.00 & 83.33 & 70.00 & 72.73 & 66.00 & 7.46 & 70.71 & 70.31 \\
\quad + RLVR
& \textbf{93.33} & \textbf{93.33} & \textbf{90.00} & 86.67 & 78.79 & 74.00 & 7.72 & 75.25 & 74.89 \\
\bottomrule
\end{tabular}
}
\end{table*}

\paragraph{Benchmarks and evaluation settings.}
For each benchmark below, we briefly describe its content, the number of samples drawn per problem, and the metric used.

\begin{itemize}
    \item \textbf{AIME 2024 / 2025 / 2026}~\citep{zhang2024aime24, aime25, aime26}: 30 problems each, drawn from the official American Invitational Mathematics Examination; competition-level mathematical reasoning. We sample 32 responses per problem and report avg@32.

    \item \textbf{HMMT 2025 / 2026}~\citep{dekoninck2026beyond}: 30 problems each from the Harvard--MIT Mathematics Tournament; high-school competition mathematics. We sample 32 responses per problem and report avg@32.

    \item \textbf{IMO-Answer}~\citep{luong2025towards}: a benchmark of International Mathematical Olympiad--style problems with verifiable closed-form final answers, used to evaluate advanced competition-style mathematical reasoning. We sample 32 responses per problem and report avg@32.

    \item \textbf{HLE}~\citep{phan2025humanity}: Humanity's Last Exam, a cross-domain benchmark of frontier-difficulty closed-form questions spanning STEM and humanities, used here as an OOD reasoning benchmark. We sample 1 response per problem and report pass@1 (equivalent to avg@1).

    \item \textbf{GPQA-Diamond}~\citep{rein2023gpqa}: 198 expert-written graduate-level questions in physics, chemistry, and biology that require multi-step scientific reasoning beyond surface lookup, used here as an OOD reasoning benchmark. We sample 4 responses per problem and report avg@4.

    \item \textbf{MMLU-Pro}~\citep{wang2024mmlu}: a multi-task benchmark of roughly 12K knowledge-intensive multiple-choice questions across 14 subject domains, used here as an OOD reasoning benchmark. Due to its scale, we sample 1 response per question and report pass@1 (equivalent to avg@1).
\end{itemize}

For benchmarks evaluated with $n>1$ samples per problem, the main table (Tab.~\ref{tab:main_results}) reports avg@$n$, the average correctness over the $n$ samples, while the appendix table (Tab.~\ref{tab:main_results_pass_at_n}) reports pass@$n$, the at-least-one-correct rate over the same samples.

\section{Pass@$n$ Per-Benchmark Results}
\label{appendix:pass_at_n}

Tab.~\ref{tab:main_results_pass_at_n} reports the per-benchmark pass@$n$ numbers for the same set of strategies as the main results table (Tab.~\ref{tab:main_results}), where the main table reports avg@$n$. Pass@$n$ measures whether at least one of $n$ samples is correct and reflects the upper-bound capability of each strategy, while avg@$n$ averages correctness across the $n$ samples. At the pass@$n$ level, post-RLVR scores saturate: top-performing methods reach 93.33 on AIME 2024/2025 and substantially compress the spread on HMMT and IMO-Answer relative to avg@$n$. Within this saturation regime, Mixed SFT still ties for or holds the top on eight of the nine benchmarks (AIME 2024/2025, HMMT 2025/2026, IMO-Answer, HLE, GPQA-Diamond, MMLU-Pro), with only AIME 2026 led by NTR and NSR. The clear advantage of Mixed SFT visible in the avg@$n$ table (Tab.~\ref{tab:main_results}) is therefore partially compressed at the pass@$n$ ceiling, because the upper-bound metric is less able to distinguish mature strategies whose best samples often agree. This further supports our choice of avg@$n$ as the primary criterion: it captures expected per-sample quality, where the differences across initialization strategies remain clearly visible.

\begin{table*}[t]
\centering
\scriptsize
\setlength{\tabcolsep}{3.5pt}
\caption{
Full per-benchmark version of the Mixed SFT ablation from Sec.~\ref{sec:analysis}. Inserting an extra next-chunk reasoning RL stage (NTR or NSR) between Mixed SFT and RLVR does not improve over directly applying RLVR on Mixed SFT.
}
\label{tab:mixed_sft_reconstruction_full}
\resizebox{\textwidth}{!}{
\begin{tabular}{lccccccccc}
\toprule
Method & AIME24 & AIME25 & AIME26 & HMMT25 & HMMT26 & IMO-Ans. & HLE & GPQA-Dia. & MMLU-Pro \\
\midrule
Mixed SFT
& 45.52 & 42.50 & 20.94 & 10.42 & 15.91 & 12.50 & 6.71 & 36.24 & 56.61 \\
\midrule
Mixed SFT $+$ RLVR
& \textbf{87.50} & \textbf{85.73} & \textbf{70.42} & \textbf{55.00} & \textbf{51.52} & \textbf{54.00} & \textbf{9.24} & \textbf{60.98} & \textbf{75.84} \\
Mixed SFT $+$ NTR  $+$ RLVR
& 86.35 & 84.58 & 70.21 & 53.73 & 50.50 & 52.25 & 7.69 & 56.44 & 74.36 \\
Mixed SFT $+$ NSR $+$ RLVR
& 87.21 & 84.67 & 69.35 & 53.15 & 49.52 & 53.26 & 8.13 & 56.36 & 74.86 \\
\bottomrule
\end{tabular}
}
\end{table*}

\section{Mixed SFT Format Instability}
\label{appendix:case_mixed_sft}

Mixed SFT jointly trains on no-CoT and long-CoT data in a single stage. The two data sources differ substantially in output structure: long-CoT data uses explicit \texttt{<think>}\ldots\texttt{</think>} markers, extended reasoning traces, and a standardized answer format, whereas no-CoT data resembles ordinary mathematical solutions with neither unified reasoning boundaries nor a fixed answer format. Exposing the model to both simultaneously creates conflicting structural conventions that the policy cannot yet resolve, temporarily destabilizing its output format even though the underlying mathematical content is largely preserved. This manifests as a pre-RLVR accuracy drop---not because the model has lost knowledge, but because the destabilized response patterns (such as multiple disjoint \texttt{<think>} segments or skipping reasoning entirely) degrade the quality of the produced solutions. RLVR's verifier reward subsequently re-imposes a consistent format, at which point the accumulated mathematical knowledge from no-CoT data is fully expressed (Fig.~\ref{fig:accuracy_format_ratio}).

\begin{figure}[t]
\centering
\includegraphics[width=\columnwidth]{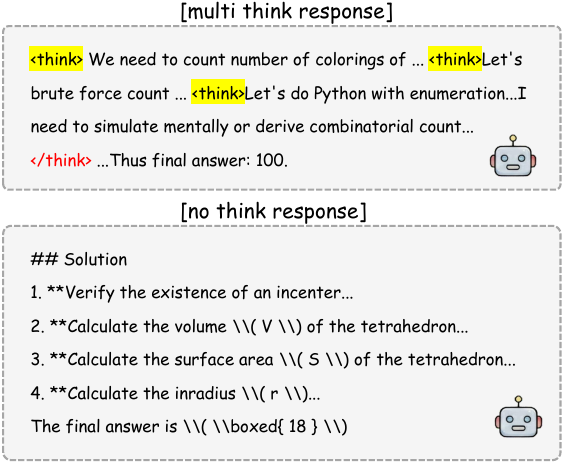}
\caption{
Two representative format-instability failure modes of Mixed SFT before RLVR: a no-think direct answer with the \texttt{<think>} block missing, and a response containing multiple \texttt{<think>} segments.
}
\label{fig:case_mixed_sft}
\end{figure}

Fig.~\ref{fig:case_mixed_sft} shows two representative failure modes observed before RLVR: \emph{(i)} a response that contains multiple \texttt{<think>} tags, and \emph{(ii)} a no-think direct answer where the \texttt{<think>} block is omitted entirely and the model jumps straight to the final answer. Both patterns disrupt the model's reasoning process and lead to lower-quality solutions, even though the underlying mathematical knowledge from no-CoT data remains intact.

\section{Full Per-Benchmark Results for the Mixed SFT Ablation}
\label{appendix:full_ablation_table}

Tab.~\ref{tab:mixed_sft_reconstruction_full} reports the full per-benchmark version of the Mixed SFT ablation in Sec.~\ref{sec:analysis}, comparing Mixed SFT followed directly by RLVR against Mixed SFT followed by an intermediate NTR or NSR stage before the same RLVR. Across all eight benchmarks, inserting an extra next-chunk reasoning RL stage after Mixed SFT yields either comparable or lower post-RLVR accuracy than directly applying RLVR on Mixed SFT, consistent with the conclusion in Sec.~\ref{sec:analysis}.

\begin{table*}[t]
\centering
\scriptsize
\setlength{\tabcolsep}{3.5pt}
\caption{
Per-benchmark post-RLVR accuracy of vanilla NTR vs.\ the entropy-controlled variant from Sec.~\ref{sec:analysis}. Both rows are followed by the same RLVR stage. Across all nine benchmarks, suppressing the entropy and length collapse \emph{lowers} the final RLVR ceiling rather than raising it.
}
\label{tab:rpt_entropy_control}
\resizebox{\textwidth}{!}{
\begin{tabular}{lcccccc|ccc}
\toprule
& \multicolumn{6}{c}{ID Reasoning} & \multicolumn{3}{c}{OOD Reasoning} \\
\cmidrule(lr){2-7} \cmidrule(lr){8-10}
Method & AIME24 & AIME25 & AIME26 & HMMT25 & HMMT26 & IMO-Ans. & HLE & GPQA-Dia. & MMLU-Pro \\
\midrule
Vanilla NTR
& 87.50 & 84.38 & 69.38 & 50.42 & 47.16 & 46.50 & 7.68 & 57.70 & 74.03 \\
Entropy-Ctrl.\ NTR
& 86.42 & 83.71 & 67.50 & 49.33 & 45.45 & 44.75 & 7.21 & 56.81 & 73.66 \\
\bottomrule
\end{tabular}
}
\end{table*}

\begin{figure*}[!tbp]
    \centering
    \includegraphics[width=0.98\textwidth]{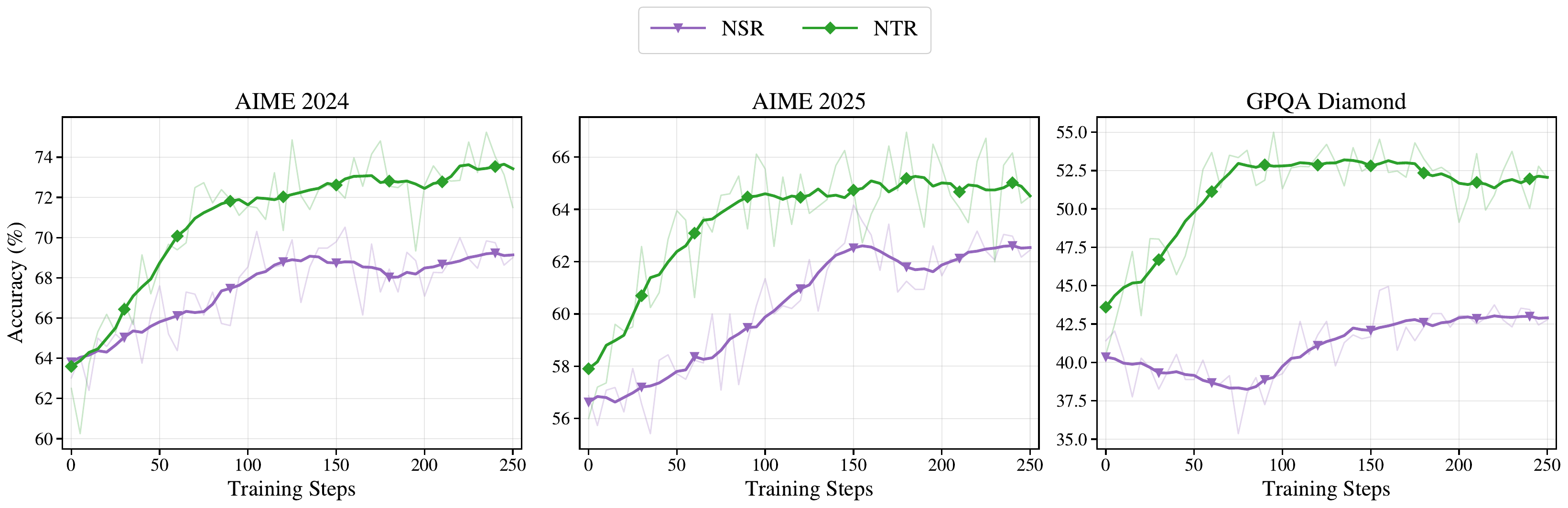}
    \caption{
    Pre-RLVR training curves of NTR and NSR on three representative benchmarks under a 32K-token evaluation context. Both stages start from the same Reasoning SFT checkpoint and are evaluated periodically up to step 250 of their respective reconstruction-RL training.
    }
    \label{fig:rpt_rlpt_curves}
\end{figure*}

\section{Entropy-Controlled NTR: Full Results}
\label{appendix:entropy_control}

Tab.~\ref{tab:rpt_entropy_control} reports the per-benchmark numbers behind the bar chart in Fig.~\ref{fig:rpt_entropy_control}. The entropy-controlled variant of NTR modifies vanilla NTR with two interventions designed to suppress entropy and length collapse. We describe both below.

\paragraph{Stochastic filtering of easy groups.}
For each rollout group, let $\rho = n_{+}/n$ denote the in-group success rate, where $n$ is the number of rollouts in the group and $n_{+}$ is the number of rollouts whose verifier reward equals $1$. The group is retained for the policy update with probability
\begin{equation}
p_{\text{keep}}(\rho) =
\begin{cases}
1, & \rho \le 0.25, \\[3pt]
1 - \dfrac{6}{5}(\rho - 0.25), & 0.25 < \rho \le 1.
\end{cases}
\end{equation}
A group is dropped from the update when an independent draw $u \sim \mathrm{Uniform}(0,1)$ exceeds $p_{\text{keep}}(\rho)$. The function equals $1$ for $\rho \le 0.25$ (groups with few successes are always kept) and then decreases linearly to $0.1$ at $\rho = 1$; it is continuous in $\rho$, so easy groups are dropped with smoothly increasing probability rather than at a hard threshold. The intent is to reduce the influence of already-easy prompts, whose remaining gradient is dominated by minor positive-advantage updates and would otherwise accelerate entropy collapse.

\paragraph{Adaptive down-weighting of positive advantages.}
Let $\bar{H}$ denote the per-token policy entropy averaged over the current batch of rollouts. For each token $t$ in rollout $i$ with raw GRPO advantage $A_{i,t}$, the entropy-controlled variant replaces $A_{i,t}$ with
\begin{equation}
\widetilde{A}_{i,t} =
\begin{cases}
0.75 \cdot A_{i,t}, & \text{if } \bar{H} < 0.5 \text{ and } A_{i,t} > 0, \\[3pt]
A_{i,t}, & \text{otherwise}.
\end{cases}
\end{equation}
When the batch becomes over-confident ($\bar{H} < 0.5$), only positive advantages are attenuated; negative advantages remain unchanged, which relatively up-weights their contribution to the policy gradient and counteracts the drift toward a single peaked template.

\paragraph{Result.}
Across all evaluated benchmarks, this entropy-controlled variant achieves \emph{lower} post-RLVR accuracy than vanilla NTR, supporting our claim in Sec.~\ref{sec:analysis} that entropy collapse is not the cause of NTR's mediocre ceiling.

\section{NTR vs NSR Pre-RLVR Training Curves}
\label{appendix:rpt_rlpt_curves}

Fig.~\ref{fig:rpt_rlpt_curves} reports the per-step evaluation accuracy of NTR and NSR during their respective pre-RLVR training stages on AIME 2024, AIME 2025, and GPQA Diamond, evaluated under a 32K-token context. Both stages start from the same Reasoning SFT checkpoint and differ only in the reconstruction objective described in Sec.~\ref{appendix:training}: NTR uses token-level prefix matching against high-entropy targets, while NSR uses a sentence-level generative judge. Starting from comparable initial accuracies on all three benchmarks, NTR improves more steeply than NSR and finishes slightly above NSR by step 250, with the gap most pronounced on GPQA Diamond. This 32K-context ordering is consistent with the 64K pre-RLVR numbers in Tab.~\ref{tab:main_results}, where NTR also slightly leads NSR on these benchmarks.

\section{Prompts}
\label{appendix:prompts}

We list below the system prompts used during NTR, NSR, and RLVR training. NTR additionally appends the no-CoT context as a user message; RLVR receives the math problem from DAPO-Math-17K as the user message; NSR receives a sentence-completion target as the user message.

\begin{figure*}[!htbp]
\centering
\includegraphics[width=0.95\textwidth]{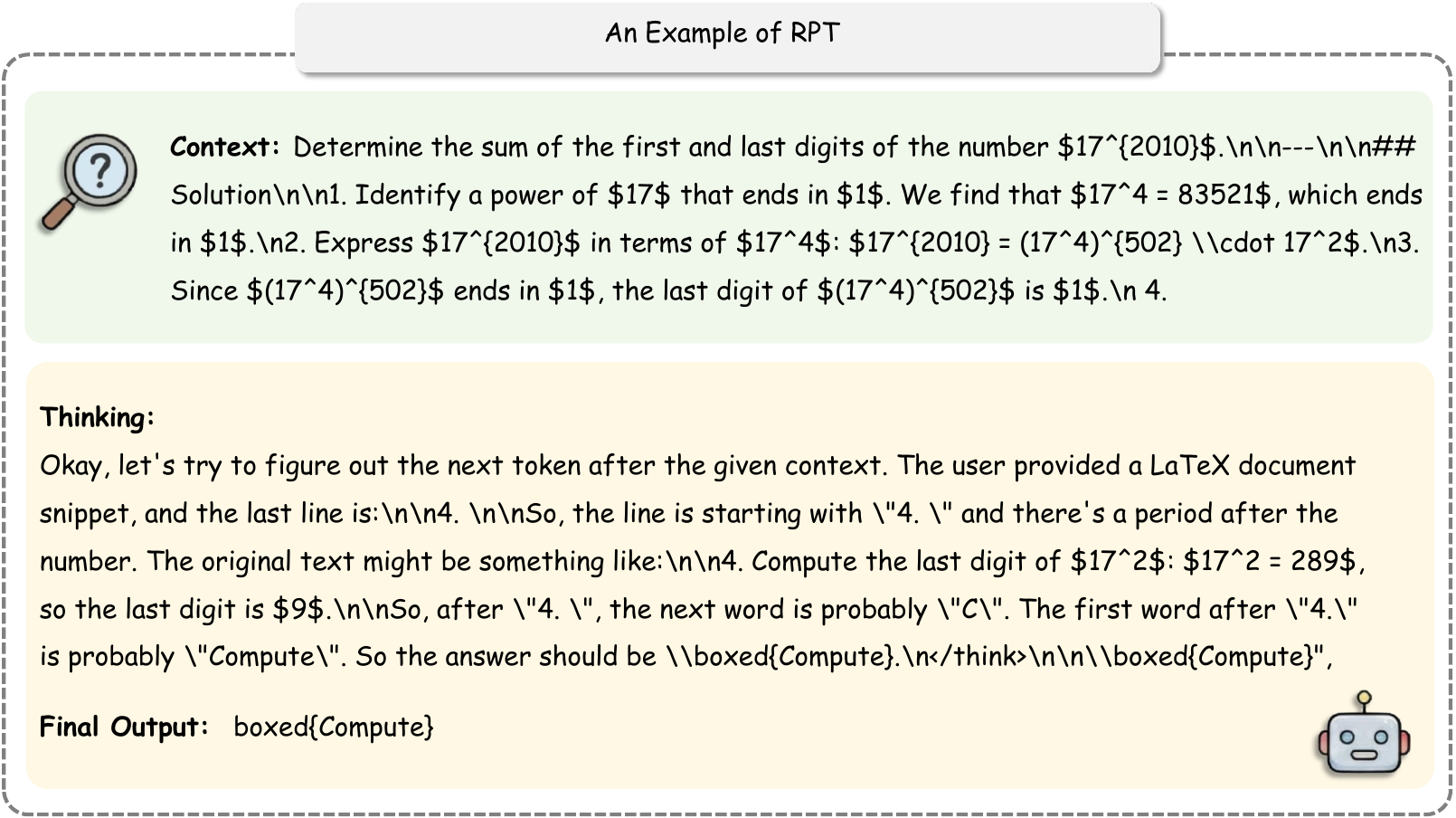}
\caption{
Representative NTR prediction cases drawn from the top-20\% highest-entropy targets of the no-CoT corpus. Even on these nominally uncertain targets, many cases can be solved by short, local-completion reasoning over a few preceding tokens rather than long-horizon inference, allowing the reconstruction reward to be satisfied without genuine reasoning.
}
\label{fig:rpt_case}
\end{figure*}

\begin{figure*}[t]
\centering
\includegraphics[width=0.95\textwidth]{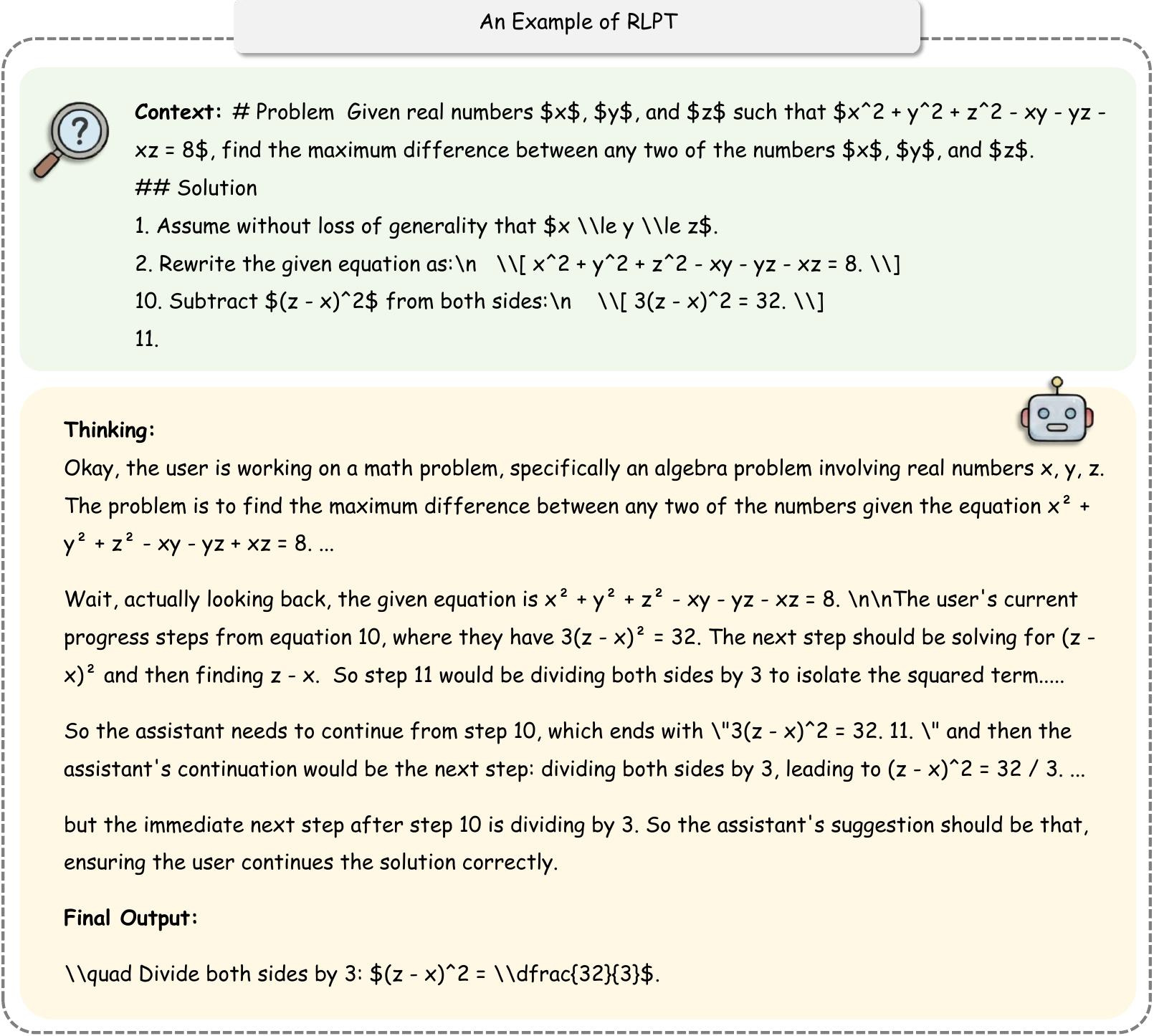}
\caption{Representative NSR training trajectory: a no-CoT prefix, the model's intermediate thinking, and the predicted next sentence judged by the generative reward model.}
\label{fig:case_rlpt}
\end{figure*}

\begin{figure*}[t]
\centering
\includegraphics[width=0.95\textwidth]{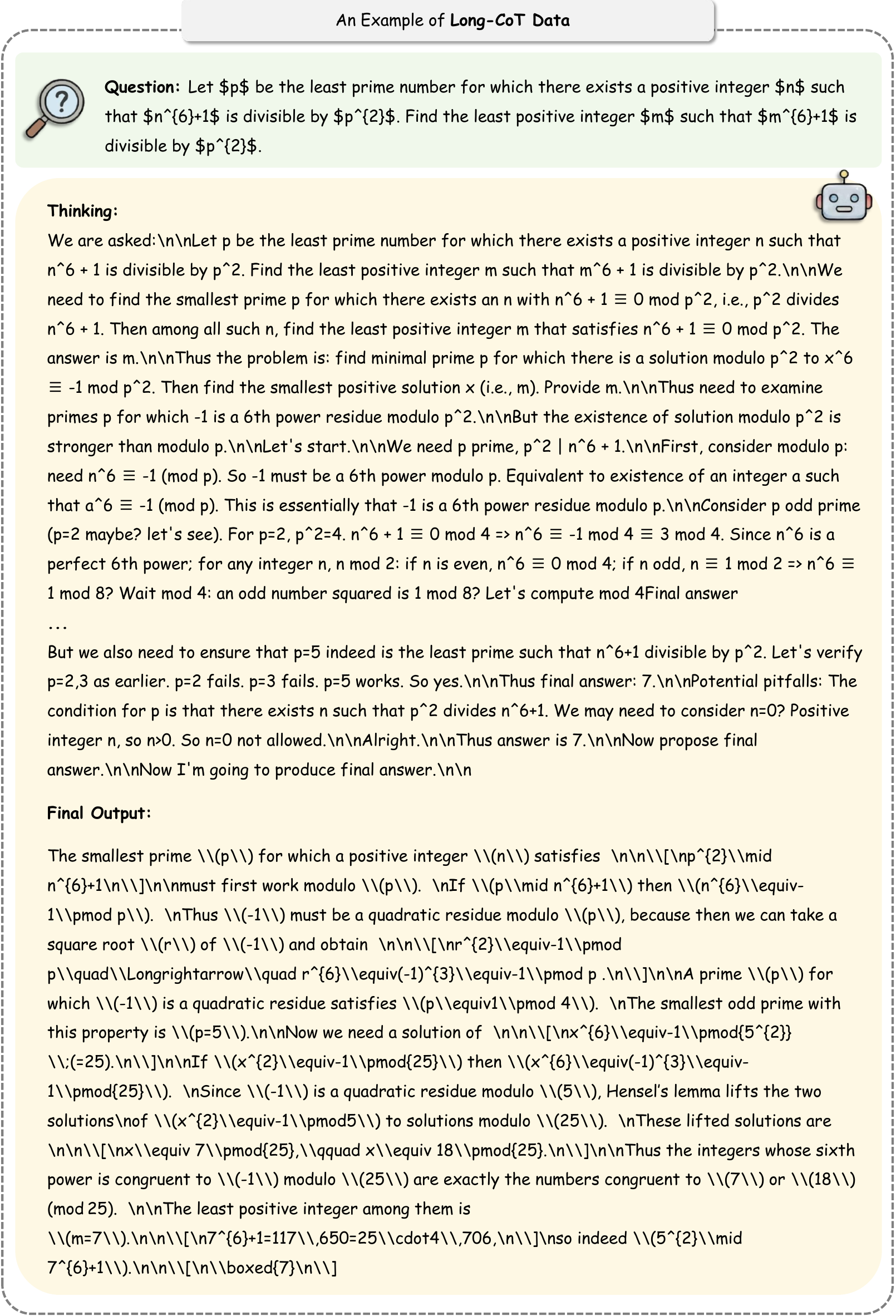}
\caption{Representative long-CoT training sample: a math problem paired with a step-by-step reasoning trajectory and final answer.}
\label{fig:case_longcot}
\end{figure*}

\begin{figure*}[t]
\centering
\includegraphics[width=0.95\textwidth]{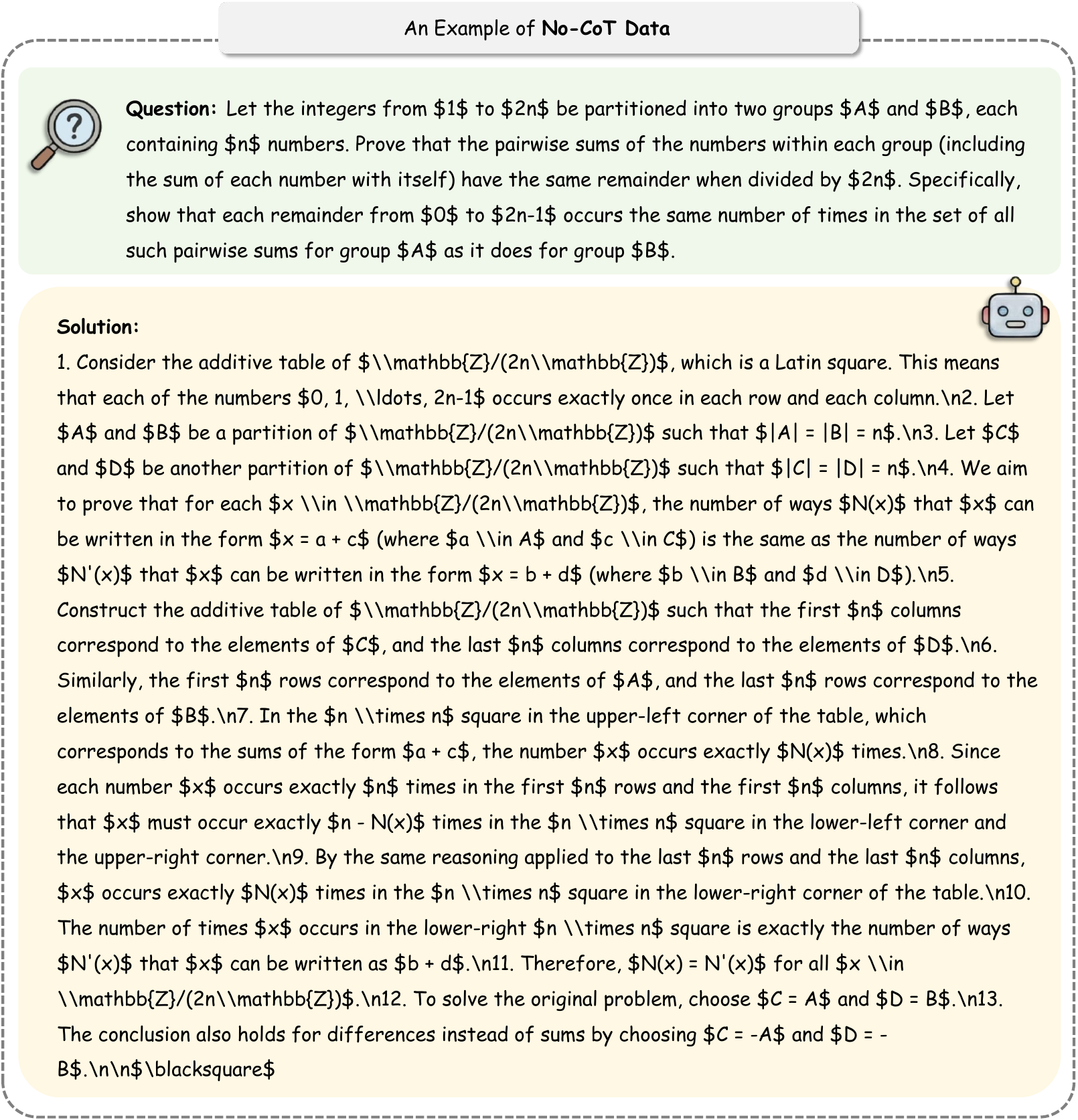}
\caption{Representative no-CoT training sample: a math problem with a compressed solution that omits explicit reasoning traces.}
\label{fig:case_nocot}
\end{figure*}

\clearpage
\onecolumn

\begin{tcolorbox}[rpt_prompt]
You are an expert reasoner with extensive experience in all areas. You approach problems through systematic thinking and rigorous reasoning. Your response should reflect deep understanding and precise logical thinking, making your solution path and reasoning clear to others. Please put your thinking process within \texttt{<think>...</think>} tags. Predict the next token and wrap it in \texttt{\textbackslash boxed\{\}} given the context. \texttt{\$Context\$}\\
\\
\texttt{\{context\}}
\end{tcolorbox}

\begin{tcolorbox}[rlpt_prompt]
You are an expert reasoner with extensive experience in completion tasks. Please put your thinking process within \texttt{<think>...</think>} tags.
\end{tcolorbox}

\begin{tcolorbox}[rlvr_prompt]
You are an expert reasoner with extensive experience in all areas. You approach problems through systematic thinking and rigorous reasoning. Your response should reflect deep understanding and precise logical thinking, making your solution path and reasoning clear to others. Please put your thinking process within \texttt{<think>...</think>} tags.
\end{tcolorbox}